\documentclass[Afour,sageh,times]{sagej}

\usepackage{moreverb,url}
\usepackage[colorlinks,bookmarksopen,bookmarksnumbered,citecolor=red,urlcolor=red]{hyperref}

\usepackage{orcidlink}
\usepackage{siunitx}
\usepackage{graphicx}
\usepackage{booktabs}
\usepackage{array}
\usepackage{caption}
\usepackage{epstopdf}
\usepackage{color, soul}
\usepackage{makecell}
\usepackage{longtable}
\usepackage{multirow}
\usepackage{gensymb}

\usepackage{rotating}
\usepackage{booktabs}
\usepackage{float}

\usepackage{enumitem}
\setlist[itemize]{left=0em, labelsep=1em} 
\setlist[enumerate]{left=0em, labelsep=0.5em} 

\usepackage{stfloats}
\usepackage{threeparttable}

\usepackage{fancyhdr}
\usepackage{etoolbox}
\usepackage{xcolor}

\graphicspath{{./FiguresInManuscript/}}

\newenvironment{orc}[1]%
{\subsection*{\normalsize\sagesf\bfseries ORCID iDs}\begin{refsize}\noindent #1}%
	{\end{refsize}}

\newcommand\BibTeX{{\rmfamily B\kern-.05em \textsc{i\kern-.025em b}\kern-.08em
T\kern-.1667em\lower.7ex\hbox{E}\kern-.125emX}}

\def\volumeyear{2016}

\newcommand{\arxivnotice}{%
	\parbox[t]{1.5\textwidth}{%
		\normalfont\small\color{gray}
		This is the accepted manuscript with post-proof corrections. It is not the Version of Record. The Version of Record has been published in \textit{The International Journal of Robotics Research} and is available at \url{https://doi.org/10.1177/02783649261444046}.
	}%
}

\begin{document}

\runninghead{K2MUSE: Multimodal Lower Limb Locomotion Dataset}

\title{%
	K2MUSE: A human lower-limb multimodal walking dataset spanning task and acquisition variability for rehabilitation robotics\\[-0.9em]
	{\arxivnotice}
}

%

\author{Jiwei Li\affilnum{1,2}~\orcidlink{0009-0000-2297-8278}, Bi Zhang\affilnum{1}~\orcidlink{0000-0001-8001-002X},
	Xiaowei Tan\affilnum{1}~\orcidlink{0000-0003-0990-0323}, Wanxin Chen\affilnum{1,2}~\orcidlink{0000-0001-5930-0209}, 
	Zhaoyuan Liu\affilnum{1,2}~\orcidlink{0009-0002-5195-6128}, Juanjuan Zhang\affilnum{3}~\orcidlink{0000-0002-3833-487X},
	Weiguang Huo\affilnum{3}~\orcidlink{0000-0002-7370-5189}, Jian Huang\affilnum{4}~\orcidlink{0000-0002-6267-8824}, 
	Lianqing Liu\affilnum{1}~\orcidlink{0000-0002-2271-5870} and Xingang Zhao\affilnum{1}~\orcidlink{0000-0001-8194-1870}}

\affiliation{\affilnum{1}State Key Laboratory of Robotics and Intelligent Systems, Shenyang Institute of Automation, Chinese Academy of Sciences, Shenyang, China\\
	\affilnum{2}University of Chinese Academy of Sciences, Beijing, China\\
	\affilnum{3}College of Artificial Intelligence, Tianjin Key Laboratory of Intelligent Robotics, Nankai University, Tianjin, China\\
	\affilnum{4}School of Artificial Intelligence and Automation, Huazhong University of Science and Technology, Wuhan, China}
	
\corrauth{Bi Zhang, State Key Laboratory of Robotics and Intelligent Systems, Shenyang Institute of Automation, Chinese Academy of Sciences, No.114 Nanta Street, Shenyang, Liaoning 110016, China.\\
Email: zhangbi@sia.cn\\
Xingang Zhao, State Key Laboratory of Robotics and Intelligent Systems, Shenyang Institute of Automation, Chinese Academy of Sciences, No.114 Nanta Street, Shenyang, Liaoning 110016, China.}
\email{zhaoxingang@sia.cn}

\begin{abstract}

The natural interaction and control performance of lower limb rehabilitation robots are closely linked to biomechanical information from various human locomotion activities. 
Multidimensional human motion data significantly deepen the understanding of the complex mechanisms governing neuromuscular alterations, thereby facilitating the development and application of rehabilitation robots in multifaceted real-world environments.
However, existing lower limb datasets are inadequate for supplying the essential multimodal data and large-scale gait samples necessary for the development of effective data-driven approaches, and the significant effects of acquisition interference in real applications are neglected.
To fill this gap, we present the K2MUSE dataset, which includes a comprehensive collection of multimodal data, comprising kinematic, kinetic, amplitude mode ultrasound (AUS), and surface electromyography (sEMG) measurements. 
The proposed dataset includes lower-limb multimodal data collected from two cohorts, including 30 able-bodied young adults and 12 older adults, across different inclines (0$^\circ$, $\pm$5$^\circ$, and $\pm$10$^\circ$), speeds (0.5 m/s, 1.0 m/s, and 1.5 m/s), and representative non-ideal acquisition conditions (muscle fatigue, electrode shifts, and interday differences).
The kinematic and ground reaction force data were collected with a Vicon motion capture system and an instrumented treadmill with embedded force plates, whereas the sEMG and AUS data of thirteen muscles on the bilateral lower limbs were synchronously recorded.
To validate the quality of the data, we quantified repeatability across locomotion modes, speeds, and inclines, examined physiological signatures under non-ideal acquisition conditions, and observed high agreement with existing public datasets. We also report baseline motion-intention recognition results, including joint angle estimation and gait phase classification, with a multimodal transformer model demonstrating accurate and stable performance. In addition, we present a control-oriented application in which an end-to-end model trained on the K2MUSE dataset provides hip assistance via a soft exoskeleton and yields consistent reductions in metabolic cost across multiple terrains. 
K2MUSE is released with the corresponding structured documentation, preprocessing pipelines, and example code, thereby providing a comprehensive resource for rehabilitation robot development, biomechanical analysis, and wearable sensing research.
The dataset is available at \href{https://k2muse.github.io/}{\textcolor{blue}{https://k2muse.github.io/}}.
\end{abstract}

\keywords{Dataset, rehabilitation robot, kinematics, kinetics, lower limb locomotion, multimodal sensing, non-ideal condition, A-mode ultrasound, surface electromyography}

\maketitle

\section{Introduction}
Rehabilitation robots, such as autonomous wearable exoskeletons and advanced prostheses, are inherently linked to human locomotion mechanisms \citep{Tan_TRO, Chen_TRO}.
In bioinspired system designs and control strategies, efficient human--robot interaction (HRI) processes rely on abundant biomechanical data derived from motor skills \citep{BioInspire1, BioInspire2, BioInspire3}.
However, acquiring well-structured human gait data presents significant challenges, as standardized procedures and reliable protocols are needed. Moreover, fully capturing the diverse range of locomotion tasks remains difficult. 
Existing motion datasets are used primarily for motion analysis and recognition \citep{DatasetReview}, revealing a gap between the available data and their applicability to human--robot coupled systems.
To meet the demands of complex human--robot coordination scenarios, the effectiveness and diversity of human motion data must be increased.

In the context of robotic system embodiment, exoskeleton controllers should be designed to operate in harmony with the human nervous system to achieve seamless interaction \citep{InternalModel}. 
To improve motor performance and restore walking ability, human motion data are often used as a reference trajectory for control strategies \citep{BioInspire2, BioInspire3}, allowing the control system to manage various rhythmic continuous movements through finite state machines \citep{Ref_Trajectory}. 
However, this control strategy is constrained by the need for predefined assistive modes, making it difficult to generalize this approach to other tasks.
In addition to finite-state tracking, thigh kinematics approaches parameterize gait progression via a phase variable for virtual constraint control, where increased phase monotonicity and linearity increase the robustness of knee--ankle trajectory tracking under variable gait conditions \citep{naeem2022virtual}.
Although human-in-the-loop optimization schemes can enhance walking assistive torques and reduce energy costs, the time burden for participants is also increased \citep{ZhangJJ, HIP_Optimization}.
To overcome these limitations, \citet{Gregg} proposed a task-adaptive controller that optimizes performance by dynamically adjusting according to specific applications and physiological states.
In addition, a different approach to designing control frameworks for multitask and multimodal motion assistance involves combining the reliability of physical modeling with the adaptability of modern artificial intelligence (AI) algorithms \citep{LiXiang}.
Further integrating the strengths of multidimensional physiological data and AI would enable exoskeletons to respond more dynamically to various conditions.

Recent advances in low-cost, heterogeneous wearable data acquisition have demonstrated that multi-sensor signals can be synchronously collected across a range of indoor and outdoor activities, enabling scalable data collection beyond strictly laboratory settings \citep{zubair2022development}.
Given the unique biomechanics of human movement, data-driven approaches provide alternative solutions to bridge the gap from laboratory technology to real-world applications, and such data-driven approaches have demonstrated potential in optimizing the effectiveness of control laws \citep{DataDriven}.
By utilizing human kinematic trajectory data from \citet{CMU_Database}, optimized exoskeleton controller parameters trained through reinforcement learning algorithms in a simulation environment can facilitate simulation-to-reality (sim2real) motion assistance across multiple scenarios \citep{Simulation}.
Data-driven methods can also be employed to establish mapping relationships between wearable sensor data and joint torques \citep{TCN_Nature}. Datasets from various walking tasks have been leveraged to estimate biological joint moments, with the resulting torques serving as control commands for the exoskeleton \citep{CyclicTasks}.
By implicitly encoding motion data, the unpredictable nature of human movement can be transformed into an end-to-end model mapping task, thereby realizing the perception and actuation loop \citep{TCN_SR}.
This approach allows the controller to effectively manage interactions among the human, robot, and environment, thereby increasing the robustness, scalability, and versatility of the control system.
Therefore, datasets containing multimodal signals and data from multiple scenarios and tasks can support the learning of end-to-end controllers that adapt to biomechanical changes in human motion. This facilitates a symbiotic relationship between humans and robots, endowing the controller with embodied intelligence \citep{SR_Editorial}.

With advancements in medicine, robotics, and AI, research on lower limb locomotion has progressively advanced. 
Datasets related to lower limb locomotion have been instrumental in related research, as numerous studies have reported the kinematics and kinetics of able-bodied participants performing steady-state activities \citep{DatasetReview}.
These datasets encompass represent a variety of tasks beyond level-ground walking \citep{Walk_PeerJ}, including running \citep{Run}, slope walking \citep{Ramp_Stair}, stair ascent/descent \citep{Stair}, and sitting and standing \citep{SitStand}.
In addition, \citet{SciDataGregg} and \citet{CyclicTasks} expanded the task spectrum by incorporating data collected under non-steady conditions and in non-cyclic tasks.
In addition to kinematic and kinetic data, signals from other modalities, such as surface electromyography (sEMG), have been explored to measure the activation of specific lower limb muscles \citep{KK_EMG1, KK_EMG2, KK_HDEMG}. 
These findings elucidate the physiological changes that occur during movement.
These datasets encompass diverse movement patterns across various terrain conditions and walking speeds.
Nonetheless, the constraints of the experimental protocols used to collect existing datasets typically limit the availability of multimodal data to a restricted number of strides \citep{KK_EMG1, KK_EMG2, KK_HDEMG, LiGuangLin}.

To increase the diversity and representativeness of datasets, a wider range of acquisition paradigms and signal modalities beyond conventional walking tasks must be explored.
Evidence from sEMG-driven prosthesis research suggests that locomotion intention can be inferred from wearable biosignals, highlighting the robustness requirements arising from terrain variations and subject-specific biomechanics \citep{hussain2020intent}.
As summarized by \citet{SeNic}, sEMG signals characterize muscle action potentials and are susceptible to various non-ideal conditions arising from different human-robot-environment interactions, such as electrode shifts, muscle fatigue, inter-day differences, and individual variability.
These conditions have been less explored in studies of lower limb motion datasets.
Investigating these non-ideal conditions is critical for implementing intention recognition technologies developed in the laboratory in real-world applications and increasing our understanding of biomechanics.
With respect to signal modality, researchers investigating muscle deformation measurement methods have explored the use of brightness-mode ultrasound to analyze parameters such as the pennation angle and its relationship with joint torque during muscle contraction \citep{BUS_Exosuit, BUS_Force}.
Despite advancements in the miniaturization of wearable devices \citep{BUS_Imaging}, hardware constraints and computational demands remain significant challenges.
Recent advancements in amplitude mode ultrasound (AUS), a lightweight technique for measuring muscle thickness, offer promising solutions for obtaining biomechanical data in addition to sEMG signals, and such methods have been widely applied in the recognition of hand and wrist motions \citep{AUS_Cyber, AUS_Online, SMG_TMECH, AUS_TIE}.
In AUS, one-dimensional scanning is employed to track muscle depth variations, thereby minimizing the dependence on image processing algorithms and complex instrumentation.
Therefore, AUS provides representations of mechanical outputs by measuring muscle deformation, thereby effectively compensating for the limitations of sEMG, which is susceptible to neurological conditions, thereby elucidating muscle physiological mechanisms.
However, its application in estimating mechanical loads during continuous lower limb movements remains less explored \citep{AUS_Torque}.
In summary, large-scale multimodal datasets that encompass a wide range of complex environments and movement patterns are lacking. This gap limits the availability of comprehensive human motion data and hinders the progression of rehabilitation robotics technology.

\begin{figure*}
	\centering
	{\includegraphics[width=1\linewidth]{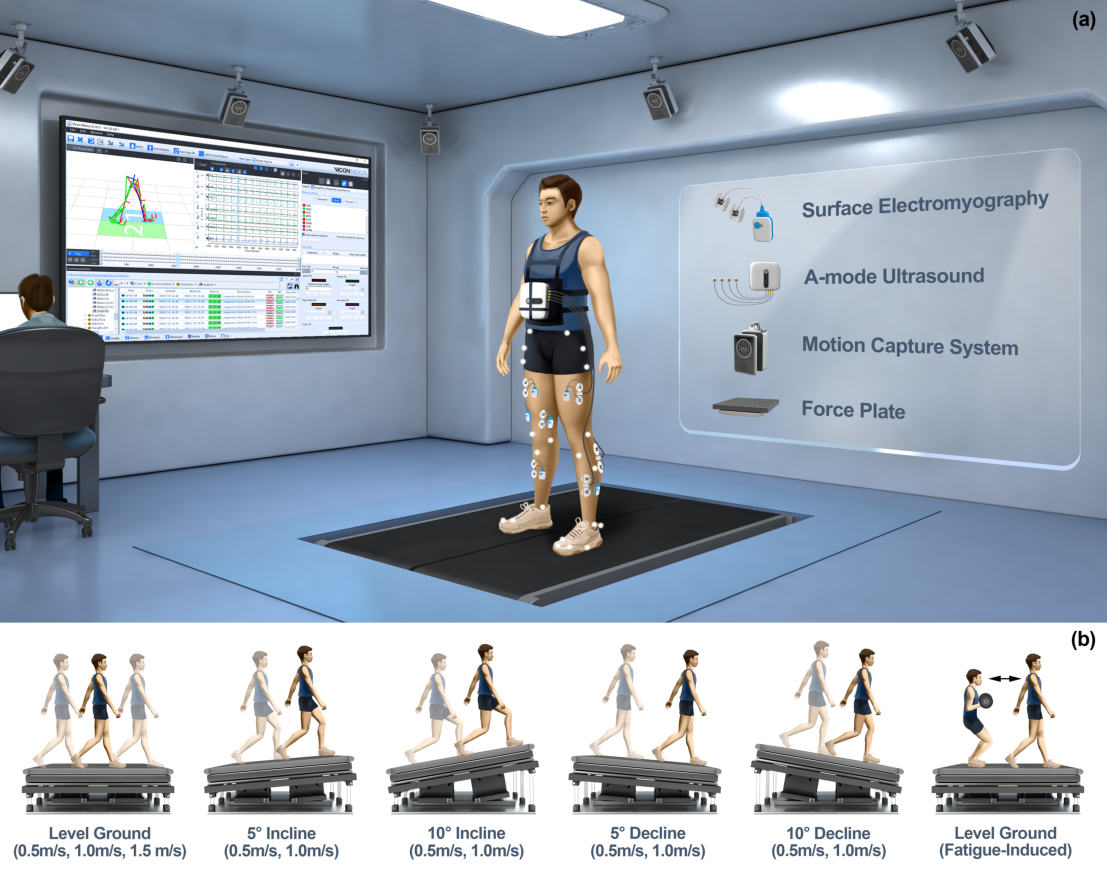}}
	\captionsetup{justification=justified, singlelinecheck=false}
	\caption{Experiments were conducted in the biomechanics laboratory. (a) The experimental scene shows a participant equipped with all the devices: a motion capture system, a treadmill with embedded force plates, an sEMG system, and an AUS device. (b) Participants performed experiments on a treadmill under diverse conditions, including different ascending and descending ramp angles and walking speeds. In the fatigue-induced experiment, the participants alternated between squatting and walking continuously.}
	\label{fig:LabScene}
\end{figure*}

In this paper, we introduce an open-source dataset, K2MUSE, which encompasses 20 ambulation conditions and includes comprehensive kinematic, kinetic, amplitude mode ultrasound and surface electromyography data, in addition to participants' anthropometric information.
The dataset was collected from 30 healthy young adults and 12 older adults. As shown in Figure~\ref{fig:LabScene}, to connect with daily living settings and encompass a wide range of locomotion modes and conditions, the data collection process involved various locomotion activities, including walking on level ground; ascending walking on 5$^\circ$ and 10$^\circ$ ramps; and descending walking on 5$^\circ$ and 10$^\circ$ ramps. 
In terms of the walking speed, three speeds were used: 0.5 m/s, 1.0 m/s, and 1.5 m/s. 
Moreover, the data collection conditions included multiple scenarios, including ideal conditions, muscle fatigue conditions, electrode shifts, and inter-day differences.
This dataset is suitable for developing control algorithms for lower limb rehabilitation exoskeletons, active prostheses, and humanoid robots, as well as for intention recognition and biomechanical analysis of lower limb movements.
K2MUSE is designed not only to encompass a broader range of walking conditions but also as a robotics-oriented multimodal locomotion resource that unifies biomechanics, physiological muscle-state sensing, and robot-task validation data within a synchronized framework. This design supports the transition from standard gait analysis to robust rehabilitation robotics applications.
The main contributions of this paper are summarized as follows:
\begin{itemize}
	\item To our knowledge, K2MUSE is the first publicly available dataset that synchronously provides sEMG, AUS, kinematic, and kinetic data, enabling joint analyses of muscle contraction dynamics and biomechanics that are difficult to conduct with conventional locomotion datasets.
	\item In addition to covering locomotion variability across multiple walking speeds and ramp inclines, K2MUSE explicitly includes representative non-ideal acquisition conditions, supporting the development of decoding models with increased robustness and generalizability to real-world clinical variability.
	\item K2MUSE is released with comprehensive technical validation and task-oriented benchmarks. These include comparisons with public datasets, baseline intention-decoding results, and representative wearable-robot use cases, providing direct and verifiable utility for rehabilitation robotics research.
\end{itemize}

\section{Related work}

High-quality lower-limb locomotion datasets are fundamental to the advancement of rehabilitation robotics and are essential for developing and benchmarking motion-intention recognition methods. Accordingly, numerous datasets have been established to capture gait mechanics across a range of walking conditions and speeds. Many of these datasets also include kinematic data in addition to wearable biosignals, such as IMUs and sEMG signals.
This section provides a summary of recent lower limb locomotion datasets, as detailed in Table~\ref{tab:RelatedWork}.

The variation in the lower limbs in three-dimensional space is a fundamental representation of lower limb ambulation, which is typically captured via inertial measurement units (IMUs) and camera-based motion capture systems \citep{IMU2020, SciDataGregg}.
In addition, when combined with kinematic data \citep{Walk_PeerJ, GaitRec}, the ground reaction force measured by force plates can be used to derive kinetic information.
\citet{IMU2020} conducted walking experiments on nine outdoor surfaces, employing six IMUs to record 3D acceleration and 3D gyroscope data from 30 participants.
To obtain joint kinetic data, \citet{Walk_PeerJ} asked volunteers to walk at self-selected speeds and eight controlled speeds on both overground and treadmill surfaces. 
However, the protocol focused mainly on steady-state gaits, which limits its ability to characterize transient or task-transition behaviors. To address this gap, \citet{SciDataGregg} extended the experimental design to include non-steady locomotion conditions.
Using a motion capture system and force plates, the experimental protocol involved participants walking and running at various speeds and inclines on a treadmill, as well as performing sit-to-stand transitions and stair ascents/descents.

\begin{sidewaystable*}[htbp]
	\centering
	\fontsize{8pt}{10pt}\selectfont
	\caption{Summary of publicly available biomechanics datasets for lower limb locomotion.}
	\label{tab:RelatedWork}
	\resizebox{\textwidth}{!}{
		\renewcommand{\arraystretch}{1} 
		\begin{threeparttable}
		\begin{tabular}{
				p{3cm}<{\raggedright} p{4cm}<{\raggedright} p{3.5cm}<{\raggedright} 
				p{0.5cm}<{\raggedright} p{0.5cm}<{\raggedright} p{0.5cm}<{\raggedright} 
				p{5cm}<{\raggedright} p{5cm}<{\raggedright} p{1.5cm}<{\raggedright}
				}
			\toprule
			\multirow{2}{*}{\textbf{Dataset}}	&	\multirow{2}{*}{\textbf{Acquisition Device}}	&	\multirow{2}{*}{\textbf{Data Modality}}	&	\multicolumn{3}{c}{\textbf{Participant Information}}	&	
			\multirow{2}{*}{\textbf{Ambulation Mode}}	&	\multirow{2}{*}{\textbf{Walking Speed}}	&	\multirow{2}{*}{\textbf{Ramp Angle}}\\
			\cmidrule{4-6}
			\textbf{}	&	\textbf{}	&	\textbf{}	&	\textbf{Total} &	\textbf{Male}	&	\textbf{Female}	&	\textbf{}	&	\textbf{}	&	\textbf{}\\
			\midrule
			\citet{Children_2008}	&	12-camera MCS, FP, sEMG sensor	&	kinematics, kinetics, sEMG	&	83	&	--	&	--	&	LG walking	&	very slow, slow, self-selected,  fast	&	N/A\\
			\midrule
			\citet{Young2011}	&	9-camera MCS, FP, sEMG sensor	&	kinematics, kinetics, sEMG	&	40	& 18	&	22	&	natural walking, toe/heel-walking, step 	&	natural speed, very slow, slow, medium, fast	&	N/A\\
			\midrule
			\citet{EEG2018}	&	ActiCap system (EEG, EOG), goniometers	&	kinematics, EEG, EOG	&	8 & 3 & 5	&	treadmill walking	&	1 mph	&	N/A\\
			\midrule
			\citet{FullBody}	&	EEG and EMG system, IMU	&	kinematics, EEG, EMG	&	10 & 5 & 5	&	LG walking, ramps,  stairs.	&	preferred speed	&	N/A\\
			\midrule
			\citet{Benchmark2018}	&	sEMG system, goniometers, IMU	&	kinematics, sEMG	&	10 & 7 & 3	&	sitting, standing, LG walking, ramp AS/DS, stair AS/DS	&	self-selected speed	&	10$^\circ$\\
			\midrule
			\citet{Walk_PeerJ}	&	12-camera MCS, FP, treadmill	&	kinematics, kinetics	&	42 & -- & --	&	LG walking	&	self-selected speed $\pm$30\%, 40--145\% of dimensionless speed	&	N/A\\
			\midrule
			\citet{KK_EMG1}	&	9-camera MCS, FP, sEMG recording system	&	kinematics, kinetics, sEMG	&	50 & 25 & 25 	&	LG walking, toe-walking, heel-walking, step AS/DS	&	nature speed, increasing/decreasing speed	&	N/A\\
			\midrule
			\citet{IMU2020}	&	IMU	&	kinematics	&	30 & 15 & 15	&	9 walking surfaces	&	 self-selected speed (16.4$\pm$4.2s per trial)	&	N/A\\
			\midrule
			\citet{SciDataGregg}	&	10-camera MCS, FPs, instrumented treadmill	&	kinematics, kinetics	&	10 & -- & 5F	&	walking, running, ramp AS/DS, sit-to-stand, stair	&	0.8--2.4 m/s	&	5$^\circ$, 10$^\circ$\\
			\midrule
			\citet{Ramp_Stair}	&	Vicon system, goniometers, IMU, FP, sEMG sensor	&	kinematics, kinetics, sEMG	&	22 & -- & -- &	LG walking, ramp, stair	&	slow, normal,  fast, 0.5--1.85 m/s	&	5.2$^\circ$ -- 18$^\circ$ \\
			\midrule
			\citet{KK_EMG2}	&	12-camera MCS, FP, sEMG system	&	kinematics, kinetics, sEMG	&	16 & 8 & 8	&	LG walking	&	1.0--4.0 $\text{km/h}$	&	N/A\\
			\midrule
			\citet{zubair2022development}	&	MyoWare (sEMG), IMU, potentiometers, flexiforce sensor	&	kinematics, sEMG, foot pressure	&	6 & 5 & 1 & LG Walking, ramp AS/DS, etc.	&	self-selected speed	&	$\pm$8$^\circ$\\
			\midrule
			\citet{LiGuangLin}	&	6-camera MCS, FPs, sEMG acquisition system	&	kinematics, kinetics, sEMG	&	40 & 30 & 10 	&	LG walking,  walking up/downstairs, etc.	&	comfortable speed	&	N/A\\
			\midrule
			\citet{IMU2023}	&	IMU, sEMG system	&	kinematics, sEMG	&	55 & 25 & 30	&	sitting, standing, walking, stair/ramp ascent/descent, etc.	&	preferred speed	&	N/A\\
			\midrule
			\citet{IMU_Ego}	&	IMU, eye tracker	&	kinematics, vision and gaze data	&	76 & 39 & 37	&	LG walking, unrestricted walking in public spaces, controlled obstacle course.	&	self-selected speed	&	N/A\\
			\midrule
			\citet{EEG2023}	&	EEG, EOG, EMG systems	&	EEG, EOG, EMG	&	14 & 9 & 5 	&	LG walking, ramp walking	&	depends on the exoskeleton 	&	N/A\\
			\midrule
			\citet{KK_HDEMG}	&	10-camera MCS, FP, IMU, EMG sensor	&	kinematics, kinetics, HDsEMG	&	10 & 5 & 5	&	LG walking, ramp walking, stair, etc.	&	slow, preferred, fast	&	5$^\circ$, 15$^\circ$\\
			\midrule
			\citet{138Adults}	&	8-camera MCS, FP, sEMG system	&	kinematics, kinetics, sEMG	&	188 & 99 & 89	&	LG walking	&	self-selected speed	&	N/A\\
			\midrule
			\citet{CyclicTasks}	&	33-camera MCS, treadmill, FPs, EMG sensor	&	kinematics, kinetics, sEMG	&	12 & 7 & 5	&	31 tasks: walking, running, jumping, etc.	&	0.6--2.5m/s	&	5$^\circ$, 10$^\circ$\\
			\midrule
			\citet{SitStand}	&	mocap system, FP, EMG, IMU	&	kinematics, kinetics, sEMG	&	65 & 23 & 42	&	Sit-to-walk	&	N/A	&	N/A\\
			\midrule
			\citet{GaitRec}	&	12-camera MCS, FP	&	kinematics, kinetics	&	20 & 9 & 11	&	LG walking	&	self-selected speed	&	N/A\\
			\midrule
			\textbf{K2MUSE (Ours)} & \textbf{8-camera MCS, FP, sEMG system, AUS device} & \textbf{kinematics, kinetics, sEMG, AUS} & \textbf{42} & \textbf{32} & \textbf{10} & \textbf{LG walking, ramp AS/DS, non-ideal conditions} & \textbf{0.5 m/s, 1.0 m/s, 1.5 m/s} &  \textbf{$\pm$5$^\circ$, $\pm$10$^\circ$}\\
			\bottomrule
		\end{tabular}
		\begin{tablenotes}
			\footnotesize  
			\item \textbf{Note:} MCS -- motion capture system; FP -- force plate/force platform; LG -- level ground; AS -- ascending; DS -- descending.
		\end{tablenotes}
		\end{threeparttable}
	}
\end{sidewaystable*}

In addition to kinematic and kinetic data, other physiological data have been included in lower limb datasets, particularly sEMG signals, which have been widely recorded in numerous datasets \citep{Benchmark2018, IMU2023}.
The study by \citet{Children_2008} addressed the gap in children's data by including multimodal gait data for level ground walking at four speeds.
Similarly, \citet{KK_EMG2} reported a dataset recorded during straight walking trials at controlled speeds ranging from 1.0 to 4.0 km/h.
In addition to walking at different speeds, the datasets proposed by \citet{Young2011} and \citet{KK_EMG1} included data collected under various other walking modes, such as toe-walking, heel-walking, and step ascending/descending.
\citet{Ramp_Stair} reported kinematic, kinetic, and sEMG data from walking trials conducted at various speeds and slopes, as well as IMUs and goniometer data.
\citet{SitStand} focused on sit-to-stand transitions and recorded kinematic, kinetic, IMU, and sEMG data from multiple joints.
\citet{KK_HDEMG} reported kinematic, kinetic, IMU, and high-density sEMG data for various tasks, such as walking at different speeds, ramp ascent/descent, sit-to-stand-to-walk, stair ascent/descent, and side-stepping gaits.
To expand the number of walking tasks included in the dataset, \citet{LiGuangLin} and \citet{CyclicTasks} designed 16 and 33 motion tasks, respectively, to capture diverse lower limb movements. These datasets include kinematic, kinetic, and sEMG data for both cyclic and non-cyclic activities.
\citet{138Adults} made significant contributions in terms of participant scale, reporting self-selected speed walking data from 138 healthy adults and 50 stroke survivors.
In addition to sEMG signals, other biological signals, such as electroencephalography (EEG) \citep{EEG2023, FullBody}, electrooculogram (EOG) \citep{EEG2018}, and egocentric data \citep{IMU_Ego}, have been recorded to capture physiological changes during walking.

However, despite substantial progress, existing lower-limb locomotion datasets often remain limited to one or more areas. Common limitations include the lack of fully synchronized kinematic and kinetic data with rich wearable physiological measurements, the lack of systematic non-ideal acquisition conditions, and an insufficient trial scale for data-driven modeling and benchmarking. To address these gaps, we release K2MUSE, a multimodal dataset that synchronously includes kinematic, kinetic, sEMG and AUS data collected across multiple walking speeds and ramp inclines, as well as under representative non-ideal conditions. 
In addition to releasing the dataset, we validate the data quality and characterize the physiological signatures under these non-ideal conditions to support studies of locomotion biomechanics and movement analysis. 
We also provide baseline evaluations, including demonstrations of intention decoding and a robotics benchmark, to facilitate the use of K2MUSE in rehabilitation robotics.
The remainder of this paper is organized as follows: Section~\ref{Methods} describes the experimental design and data acquisition procedures. Section~\ref{DatasetStructure} introduces the dataset structure and documentation. Section~\ref{AnalysisandValidation} presents the technical validation and benchmark analyses. In Section~\ref{PotentialApplications}, potential applications are discussed. Sections \ref{UsageNotes} and \ref{CodeAvailability} provide usage notes and code availability, respectively.

\section{Methods}
\label{Methods}

\subsection{Participants}

The dataset was collected from 30 healthy young adults and 12 older adults. The healthy cohort included 20 males and 10 females. Their ages ranged from 23 to 35 years (26.97$\pm$2.39 years), their heights ranged from 145.2 to 186.7 cm (172.95$\pm$9.42 cm), and their body masses ranged from 42.65 to 113.0 kg (69.65$\pm$15.04 kg). 
The older cohort included 12 male participants, all of whom were workers. Their ages ranged from 45 to 60 years (53.08$\pm$5.23 years), their heights ranged from 166 to 178 cm (172.00$\pm$4.31 cm), and their body masses ranged from 66.5 to 98.5 kg (79.69$\pm$11.06 kg).
All participants were recruited from the Shenyang Institute of Automation, Chinese Academy of Sciences.

None of the healthy participants reported any neurological diseases or musculoskeletal dysfunctions that could affect lower extremity motor performance. 
Some of the older participants reported a history of prior injuries. Detailed information is provided in the `ParticipantInformation.xlsx' file.
The experimental protocol was approved by the Ethics Committees of the Shenyang Institute of Automation, Chinese Academy of Sciences (Approval No.: 2024-03), and the People's Hospital of Liaoning Province, China (Approval No.: 2023MSLH-117) and was conducted in accordance with the Declaration of Helsinki.
All the participants were briefed about the procedures and potential risks and provided written informed consent before they participated in the experiment.

\subsection{Instrumentation}
Data were collected at the Biomechanics Laboratory at the Shenyang Institute of Automation, Chinese Academy of Sciences.
The laboratory is equipped with a motion capture system, an instrumented treadmill, an sEMG acquisition system, and an amplitude mode ultrasound acquisition system to collect kinematic, kinetic, and biological signals as participants performed various movements.
Data synchronization between devices was achieved through the square wave voltage signal generated by the Vicon Lock Lab.
The placement of retro-reflective markers and sensors was performed by two experienced assessors. To minimize inter-operator variability, the final positioning was confirmed through cross-verification to ensure precise alignment with anatomical landmarks.

\subsubsection{Treadmill.}
All walking trials were conducted on a fully instrumented treadmill (\href{https://www.bertec.com/}{Bertec, Columbus, Ohio}).
The treadmill features two independently controllable belts, each equipped with a force plate underneath, to capture six-component force data at 1000 Hz.
The treadmill incline was configured via software, and its start/stop functions and speed were controlled by custom MATLAB code. 
The treadmill acceleration was consistently set to 0.25 $\text{m/s}^2$ in all the acquisition experiments.

\subsubsection{Motion capture.}
Three-dimensional kinematic trajectories of the retro-reflective markers (14 mm diameter) were recorded at 100 Hz via a motion capture system consisting of eight Vicon V5 cameras (\href{https://www.vicon.com/}{Vicon, Oxford, UK}).
Before the formal experiments commenced, motion capture volume calibration of the cameras was conducted via a Vicon Active Wand.
The Plug-in Gait model of Vicon was utilized for kinematic and kinetic modeling.
Joint angles were calculated on the basis of the positions of the markers. The joint forces, moments, and powers were subsequently derived on the basis of the kinetic data of the force plates and the kinematic data of the markers via the Plug-in Gait Dynamic operation in the Nexus software.
A comprehensive description of the Plug-in Gait model, including the specifications of the kinematic and kinetic calculations, can be found in the Vicon Nexus User Guide \citep{NexusUserGuide} and the Plug-in Gait Reference Guide \citep{PlugInGaitGuide}.
The placement of the markers followed the Vicon Plug-in Gait lower body model marker set, with an additional set of markers attached to the participants to improve the fault tolerance of motion capture.
These additional markers were used to fill in the missing data frames in the marker trajectories after the acquisition trials.
The names and locations of all the markers are shown in Figure~\ref{fig:MarkerSet}, and detailed descriptions of specific marker placements are provided in the Plug-in Gait Reference Guide.

\subsubsection{Electromyography.}
A wireless EMG sensor system (\href{https://www.noraxon.com/}{Ultium EMG, Noraxon, USA}) was used to record the muscle activity of the bilateral lower limbs at a sampling rate of 2000 Hz.
In the hardware setup of the MR software (version 3.16, Noraxon, USA), the high-pass and low-pass filter frequencies were set to 20 Hz and 500 Hz, respectively.
The Ultium EMG system was connected to the Vicon Lock Lab via a synchronization system (MyoSync, Noraxon, USA) to receive the synchronous square wave signal.
The surface EMG sensors and Ag/AgCl electrodes were affixed to the skin via double-sided tape.
The electrode attachment positions were determined through repeated palpation, following the SENIAM recommendations \citep{SENIAM}.
As shown in Figure~\ref{fig:MuscleSet}, the sEMG signals were recorded bilaterally from different muscles, which are highlighted in blue font. 
On the right side, data from 9 muscles were captured: the tibialis anterior (TA), medial gastrocnemius (MG), lateral gastrocnemius (LG), soleus (SOL), rectus femoris (RF), vastus lateralis (VL), vastus medialis (VM), biceps femoris (BF), and semitendinosus (SEM) muscles.
On the left side, data from 4 muscles, TA, LG, RF, and BF, corresponding to the channels of the amplitude mode ultrasound device were recorded.

\subsubsection{Amplitude mode ultrasound.}
\begin{figure*}
	\centering
	{\includegraphics[scale=0.9]{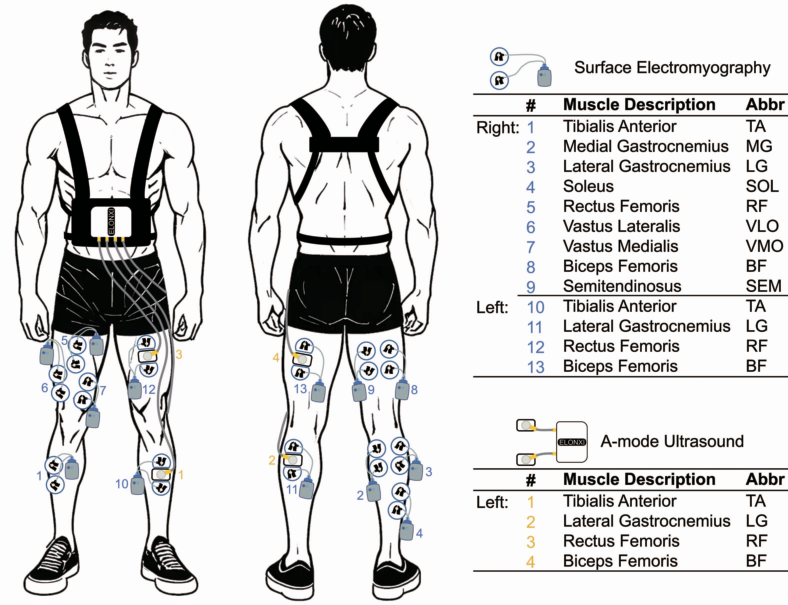}}
	\captionsetup{justification=justified, singlelinecheck=false}
	\caption{The sEMG sensors and AUS transducers were attached to the participants' skin. The channels of different instrumentation are highlighted in different colors for easy distinction. The symbol `\textbf{\#}' corresponds to the channel numbers of the different devices.}
	\label{fig:MuscleSet}
\end{figure*}

A wireless commercial four-channel amplitude mode ultrasound (AUS) device (\href{http://www.elonxi.cn}{ELONXI, China}) was used to record AUS signals at a sampling frequency of 20 Hz, with each frame containing 1000 samples \citep{AUS_Cyber}.
The frequency of the transducers was 5 MHz (9 mm diameter, 11 mm height), allowing a detection depth of up to 38.5 mm, which is sufficient for measuring contraction and extension changes within muscles.
To ensure synchronous operation with the other devices, the AUS device was connected to the Vicon Lock Lab via a synchronization line to receive the trigger voltage signal.
The AUS transducers were coated with an appropriate amount of standard ultrasound coupling agent and then secured to the skin surface of the left leg with PU film medical tape.
These transducers corresponded to the TA, LG, RF, and BF muscles, as indicated in yellow in Figure~\ref{fig:MuscleSet}.

\begin{figure*}
	\centering
	{\includegraphics[scale=0.9]{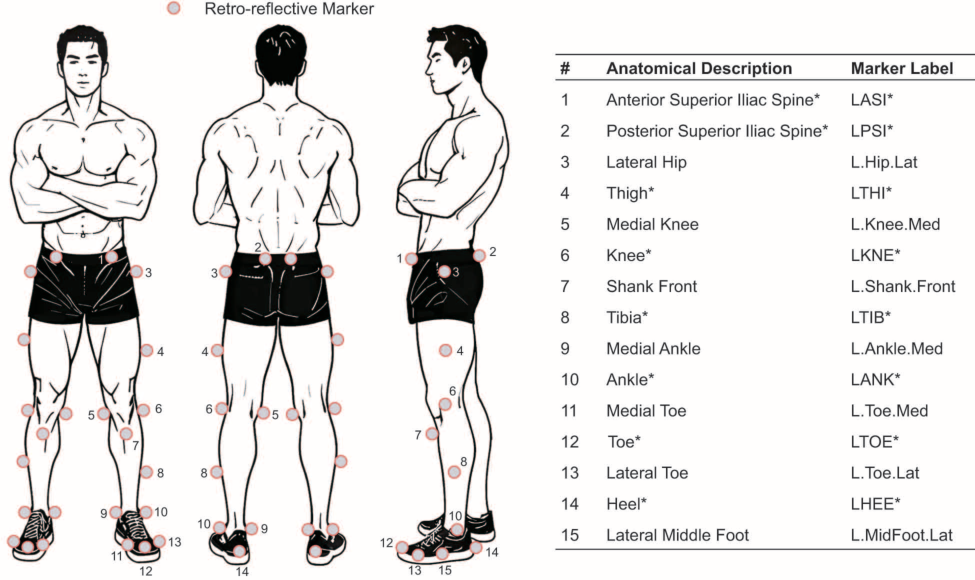}}
	\captionsetup{justification=justified, singlelinecheck=false}
	\caption{The modified marker set for motion capture. The markers were attached to the lower limbs in a generally symmetrical arrangement, with the markers on the left side shown. The markers marked with `*' were defined according to the Plug-in Gait lower body model, which implements the Conventional Gait Model. Detailed marker placement instructions for the Plug-in Gait lower body model can be found in the Plug-in Gait Reference Guide.}
	\label{fig:MarkerSet}
\end{figure*}

\subsection{Experimental protocol}
\label{Experimental protocol}
In this study, kinematic, kinetic, AUS and sEMG data were captured during ambulation at various speeds and inclines under a range of acquisition conditions. All ambulation modes were performed on an instrumented treadmill.
A custom MATLAB code was used to control the synchronous start and stop of all the device acquisitions.
Before each participant arrived, the eight Vicon cameras of the motion capture system were calibrated following the instructions provided in the Vicon Nexus User Guide. 
The force plates on the Bertec treadmill were hardware reset and zeroed in the Nexus software. 
The force plates were recalibrated to zero after the treadmill incline was adjusted.

\subsubsection{Participant preparation.}
After providing informed consent and receiving a brief introduction to all the trials, the participants were asked to wear tight-fitting shorts and appropriate athletic shoes to ensure that the transmalleolar axis was exposed.
The participant's top was subsequently secured and tightened with an elastic strap to prevent clothing from obscuring the markers on the pelvis while walking.
Before the dynamic capture trials were performed, the following steps were carried out.
\begin{enumerate}
	\item The following anthropometric measurements were taken for the Plug-in Gait lower body model: body mass, height, leg length, knee width and ankle width. The measurements were used as inputs for the lower body model. Comprehensive information about all individuals can be found in the `ParticipantInformation.xlsx' file.
	\item Before the markers and sensors were placed, the participant's hair was removed from the recording sites, and cotton pads containing 75\% alcohol were used to clean sweat, keratin, and oil off the skin surfaces.
	\item Sixteen retro-reflective markers were attached to the skin above the anatomical landmarks, and fourteen additional markers were placed on the lower body through manual palpation. All the markers were securely fixed to the skin using a combination of double-sided tape and PU film medical tape at their base.
	\item Thirteen pairs of Ag/AgCl electrodes were attached to muscle locations, and sEMG sensors were placed next to the electrodes using double-sided adhesive tape. The electrodes and sEMG sensors were connected via specially designed electrode wires. PU film medical tape was used to reinforce the electrodes and sensors. 
	\item Four transducers coated with a coupling agent were placed between paired electrodes, which were trimmed properly to accommodate the transducers. The transducers were connected to the hardware ultrasound system, which was secured to an adjustable vest using elastic bands and hook-and-loop fasteners. The cables were wrapped with non-woven fabric tape to prevent them from covering the markers during locomotion.
\end{enumerate}

After the preparations were completed, a labeling skeleton was created for the participant with the Nexus software, and a static calibration was performed to tailor the skeleton to the participant.
The participants were instructed to stand comfortably in the middle of the treadmill with their feet shoulder-width apart and positioned on either side of the belt. They maintained an upright posture with their arms crossed in front of their chest to ensure that all the markers were visible. A static trial was then performed, and the calibration was finalized using a frame where all the markers were fully visible.
The collection of dynamic locomotion data was subsequently initiated.

\subsubsection{Ideal conditions.}
All the participants in the dataset were involved in the ideal condition experiment.
The treadmill inclinations included level ground (LG), descending ramp (DS), and ascending ramp (AS), with angles of 0$^\circ$, $\pm$5$^\circ$, and $\pm$10$^\circ$, respectively. 
The transition between ramps was achieved by altering the participant's orientation on the treadmill, with positive angles for ascending and negative angles for descending.
The participants first walked on level ground at speeds of 0.5 m/s, 1.0 m/s, and 1.5 m/s.
The treadmill ramp was subsequently adjusted to 5$^\circ$, and participants walked at -5$^\circ$ and +5$^\circ$ inclines at speeds of 0.5 m/s and 1.0 m/s for both ramps. 
Finally, the treadmill incline was adjusted to 10$^\circ$, and participants walked at -10$^\circ$ and +10$^\circ$ inclines at the same speeds as those used at 5$^\circ$.
For each ramp, the walking tests at each speed were repeated five times, resulting in a total of 55 trials.

\subsubsection{Muscle fatigue.}

The muscle fatigue protocol included two components. The first component was a progressive fatigue experiment in which fatigue was gradually induced through lower-limb exercises. The second component was a sustained fatigue trial in which participants walked continuously under uphill conditions to elicit and maintain fatigue. A progressive fatigue experiment was performed after all tests under ideal conditions were completed. Participants were instructed to rest for 5 min before the fatigue protocol began. The sustained fatigue trial was conducted on a separate day.

\begin{figure*}[htbp]
	\centering
	{\includegraphics[scale=0.9]{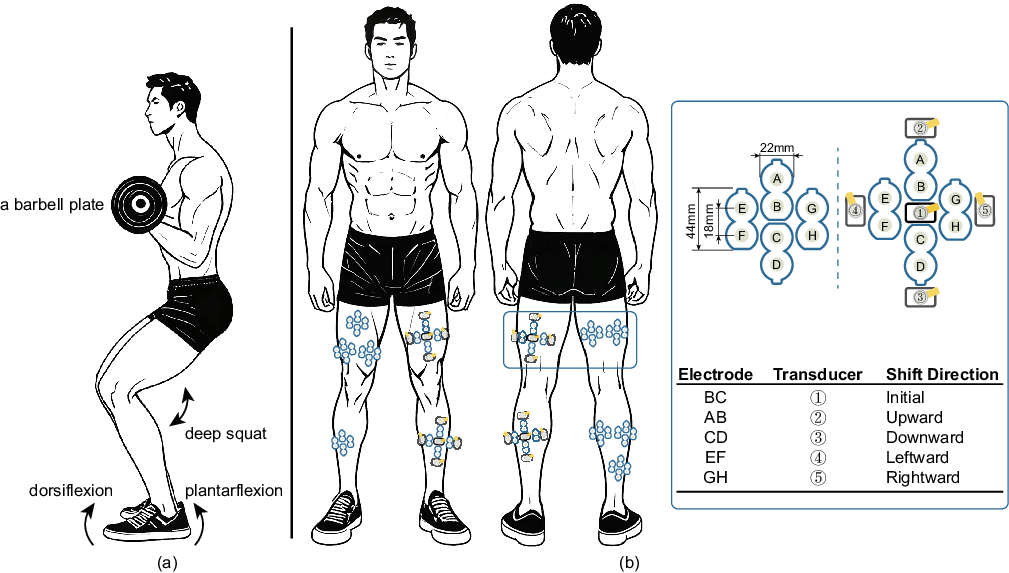}}
	\captionsetup{justification=justified, singlelinecheck=false}
	\caption{Experimental setups for muscle fatigue and electrode shift experiments. (a) Fatigue-induced lower limb exercises, including dorsiflexion/plantar flexion and squats, with a barbell held in the hands. (b) Experimental setup simulating electrode shifts, where different electrode pairs and transducers correspond to the initial positions and four shift directions.}
	\label{fig:FatigueAndShift}
\end{figure*}

\begin{itemize}
	\item \textbf{Component 1:} In the progressive fatigue experiment, the participants were instructed to walk on level ground at a speed of 1.0 m/s. Between each trial, the participants performed a set of lower limb exercises to induce muscle fatigue, as shown in Figure~\ref{fig:FatigueAndShift}~(a).
	The participants stood naturally with their feet shoulder-width apart on the ground and held a barbell plate (10 kg for males, 5 kg for females) with both hands. They then performed a complete lower limb exercise, including one repetition of standing plantar flexion and dorsiflexion, followed by one deep squat.
	After completing 10 full lower limb exercises, the participants immediately returned to the treadmill for the next walking trial without rest. The process was repeated until the participant completed 10 walking trials.
	\item \textbf{Component 2:} For the sustained fatigue experiment, participants were instructed to walk continuously for 6 min on an ascending ramp with a 10$^\circ$ incline. In addition to sEMG, AUS, kinematic, and kinetic measurements, we collected respiratory gas exchange data to assess the physiological validity of fatigue progression. To collect these data, participants wore an indirect calorimetry device (Oxycon Mobile, Vyaire Medical, Germany) throughout the trial. Metabolic cost was then computed from the recorded gas data and used as an indirect indicator of whether participants became fatigued during the protocol. The indirect calorimetry device is shown in Figure~\ref{fig:ControlFramework}, and the computation of metabolic cost followed \citet{brockway1987derivation}.
\end{itemize}

\subsubsection{Electrode shift.}

The electrode shift experiment comprised two components. The first component involved two-dimensional in-plane shifts, and the second component focused on variations along the depth direction. Eight participants performed each experiment. The two experiments were conducted on separate days. In each session, participants walked on level ground at a speed of 1.0 m/s.

\begin{itemize}
	\item \textbf{Component 1:} As illustrated in Figure~\ref{fig:FatigueAndShift}(b), to simulate electrode shifts in the two dimensional plane, an electrode constellation was attached around the muscle bellies.
	This constellation consisted of four symmetrically arranged bipolar electrodes. The initial electrode position was determined by the difference between B and C, which corresponded to the nominal electrode placement. The shift locations are labeled AB, CD, EF, and GH, corresponding to upward, downward, leftward, and rightward offsets, respectively. The numbers 1 through 5 indicate the positions of the transducers, with their shift directions aligned to match those of the electrodes.
	The participants first completed the preparation steps and performed walking trials with the initial position.
	The assessors then manually adjusted the positions of the differential electrodes and transducers in the four directions sequentially and repeated the experimental procedure with the initial position. This process continued until five trials were completed for each shift position. 
	For each participant, a total of 25 trials were recorded under different two-dimensional plane electrode shift locations.
	\item \textbf{Component 2:} Changes along the depth direction were simulated by modulating the contact pressure between the electrodes and the skin under two conditions, termed superficial and compressed.
	In the superficial condition, the electrodes and transducers were placed according to the previously defined initial position and adhered to the skin surface, representing the baseline contact pressure. In the compressed condition, an elastic athletic bandage was tightened to secure all the electrodes and transducers, thereby increasing the contact pressure. For each participant, a total of ten trials were recorded to capture signal variations induced by different pressure levels.
\end{itemize}

\subsubsection{Inter-day difference.}

Eight participants completed this experiment, which was conducted on a separate day using the same level-ground walking protocol as under the ideal condition at speeds of 0.5, 1.0, and 1.5 m/s. The electrode and sensor placements and the overall procedure were identical to those of the ideal-condition setup, with no electrode-shift or fatigue-induction procedures. Each participant completed 15 trials in the inter-day difference experiment. For participants who completed multiple sessions, the ideal-condition data and the ``initial position'' recordings from the electrode-shift experiment were also included in the inter-day difference subset.

\subsection{Synchronization procedure}
For all the conditions, the data collection procedure was the same across all the trials. All the codes and software were run on the same computer (DELL Precision 7920).
Before the start of each trial, the participants were instructed to stand naturally in the middle of the treadmill. The operator modified the values of the variables in the MATLAB code, including the acquisition condition, ambulation mode, treadmill ramp, speed, and trial number (e.g., `Ideal', `LG', `0', `1.0', `1').
The MATLAB code sent the start messages to the Nexus software via UDP communication to begin data collection, with the Nexus software prompted to set the Vicon Lock Lab ports to high voltage (5 V). 
At this point, the sEMG and AUS devices connected to the Vicon Lock Lab were synchronously triggered to start data acquisition, whereas the MATLAB code established a TCP communication connection with the treadmill.
Next, the MATLAB code generated five short beeps, each lasting 1 second, followed by one long beep to signal the participant to prepare for the beginning of the experiment. The treadmill was then started according to the speed and ramp parameters.
After the participant walked for 40 seconds, the MATLAB code sent stop messages to the Nexus software, causing the Vicon Lock Lab ports to switch to low voltage (0 V). Consequently, all the devices stopped collecting data synchronously, and the treadmill ceased operation. 
Therefore, in each trial, data were collected for 45 seconds. 
In addition to the muscle fatigue experiment, the participants were instructed to rest between trials for each condition to avoid fatigue.

\subsection{Data Elaboration}

\subsubsection{Motion capture processing.}
After data acquisition, motion capture data were processed via Vicon Nexus software (version 2.10, Vicon, Oxford, UK). 
Retro-reflective markers were labeled, and their 3D coordinates were reconstructed. This process allowed the assignment of specified marker names, as illustrated in Figure~\ref{fig:MarkerSet}.
Trajectory gaps caused by visual occlusions were filled using the recommended gap-filling workflow in the Vicon Nexus software. Spline fill was applied only to short gaps, with a maximum length of 20 consecutive frames. Longer gaps were not repaired using spline interpolation. Instead, rigid-body fill was preferred on the basis of the remaining markers for the same segment, and pattern fill or kinematic fill operations were applied when appropriate. Unlabeled marker trajectories were removed. Kinematic and kinetic variables were then computed by running the Plug-in Gait Dynamic pipeline in Vicon Nexus.
The marker trajectories were filtered via a $\text{4}^\text{th}$, zero-lag Butterworth low-pass filter with a cutoff frequency of 6 Hz and a Woltring filter with the MSE mode and a smoothing parameter of 20.
The force plate data were filtered via the same Butterworth low-pass filter applied to the marker trajectories.
Gait cycle events, including heel strike (HS) and toe-off (TO) events, on both sides of the body were identified on the basis of vertical ground reaction forces (threshold: 20 N) collected by the force plates and the trajectories of the foot markers.
The dynamic Plug-in Gait model was subsequently used to calculate the joint kinematics and kinetics by integrating the motion capture data with participant-specific anthropometric characteristics.
Finally, the processed data were exported as `*.c3d' files, which included joint angles, joint forces, joint moments, joint power, and gait events.
Detailed information on the processing of the motion capture data is available in the Vicon Nexus User Guide and the Plug-in Gait Reference Guide.

\subsubsection{Data processing.}
The sEMG data for all participants were stored in the MR software database and exported as `*.mat' files, whereas the AUS data were directly saved as `*.txt' files upon completion of data acquisition.
To parse the multimodal data into strides and perform time normalization, the participants' data were compiled into a unified MATLAB structure. Further postprocessing was conducted in MATLAB (version 2022b, MathWorks, Natick, MA, USA).
A stride was defined as the motion cycle between two consecutive heel strikes of the same leg.
For the walking trials, the MATLAB pipeline was used to extract joint angles, joint forces, joint moments, joint power, and gait events from the `*.c3d' files, and the data were parsed into strides on the basis of the identified gait events.
Each stride was subsequently time normalized by linearly interpolating the data sequence to 200 data points, ensuring consistent sampling for each gait cycle. To retain steady walking gait cycles for each trial, the data collected during the treadmill startup period were excluded on the basis of the treadmill speed, acceleration, and standing time for each trial. Only gait cycles from steady walking periods were parsed and time normalized.
Specifically, the exclusion duration for each trial was determined as follows:
\begin{equation}
	T_{\text{exclude}} = T_{\text{prepare}} + \frac{v_{\text{target}}}{a}.
\end{equation}
Here, $T_{\text{exclude}}$ is the total time excluded at the beginning of the trial, $T_{\text{prepare}}$ denotes the preparation period before belt acceleration, ${v_{\text{target}}}$ is the target walking speed, and $a$ is the constant treadmill acceleration.

Afterward, the ultrasound data from the four AUS channels for each trial were extracted and merged on the basis of the corresponding `*.txt' files. The ultrasound frames were segmented into cycles on the basis of the gait events. 
Similarly, the sEMG data were extracted from individual channels in the `*.mat' files and segmented according to the corresponding gait events on each side of the body.
Finally, the processed sEMG and AUS data were merged into the unified MATLAB structure.

\section{Dataset structure}
\label{DatasetStructure}

All the data are uploaded to \href{https://kaggle.com/datasets/98d67c253a7c820668aed0690cae20343481b8f8f8e0dafbe93b0c76d91f0ce6}{\textcolor{blue}{Kaggle}}, and detailed descriptions are available at \href{https://k2muse.github.io/datasets/}{\textcolor{blue}{https://k2muse.github.io/datasets/}}.
The K2MUSE dataset comprises kinematic, kinetic, ultrasound, and sEMG data, along with the anthropometric information of the participants, as illustrated in Figure~\ref{fig:DatasetStructure}~(a).
The participants' basic information is stored in a file named `ParticipantInformation.xlsx', which includes details such as the participants' ID number, gender, birthday, age, height, weight, shoe size, knee width, ankle width, and leg length.
The `figData.mat' file contains the average joint angles, joint moments, and joint power values for all participants, which were used to plot the curves shown in Figure~\ref{fig:IdealKinematicKinetic}.

\subsection{Processed data}
\begin{figure*}
	\centering
	{\includegraphics[width=1\linewidth]{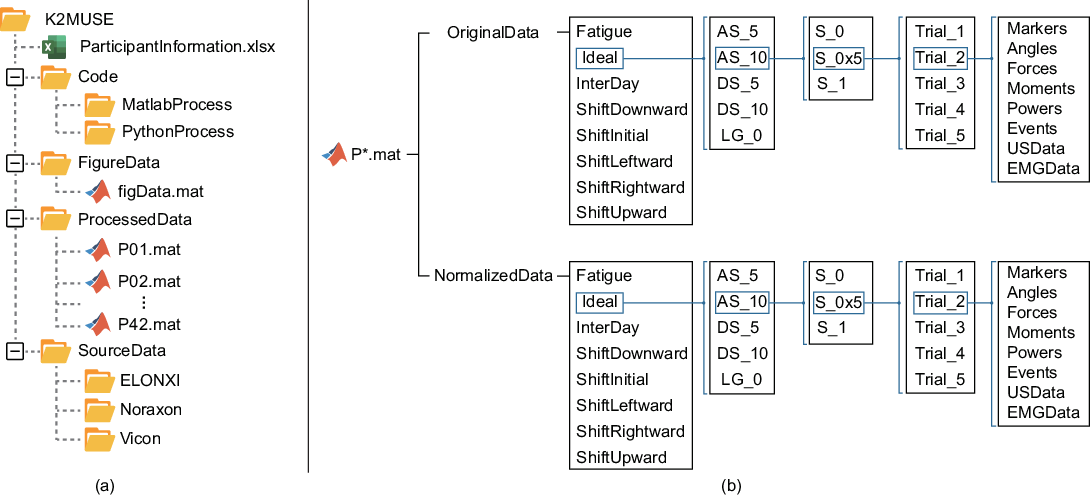}}
	\caption{Data organization outlines (a) the folder structure of the dataset and (b) the structured 'P*.mat' file for the participant.}
	\label{fig:DatasetStructure}
\end{figure*}

To facilitate the use of the dataset in this study, all the modality data are stored in the `\textit{ProcessedData}' folder. 
Each participant's data are organized into a unified MATLAB structure and saved in the format `P*.mat', where `*' corresponds to the participant ID (ranging from 1--42). 
Each `P*.mat' file follows a standardized field structure, as shown in Figure~\ref{fig:DatasetStructure}~(b).
The `P*.mat' files include kinematic data (3D marker trajectories, and 3D joint angles), kinetic data (3D joint forces, 3D joint moments, and 3D joint power), AUS data, sEMG data, and gait events.
The `OriginalData' field corresponds to the continuous data for each trial, which were directly extracted from the files exported by different devices without any further processing.
The `NormalizedData' field contains the same data but normalized by stride and segmented into gait cycles, ensuring consistency and facilitating analysis.
The naming convention for the substructures of the `OriginalData' and `NormalizedData' fields is as follows: (condition).(ambulation mode).(speed).(trial).(datatype). 
This structure ensures that the data are clearly categorized on the basis of the experimental conditions, ambulation modes, walking speed, trial number, and specific datatype.
The data structure and descriptions stored in the trial fields of `OriginalData' and `NormalizedData' are presented in Table~\ref{tab:OriginalData} and Table~\ref{tab:NormalizedData}, respectively.
Further detailed descriptions of the `P*.mat' files in the dataset are available at \href{https://k2muse.github.io/datasets/}{\textcolor{blue}{https://k2muse.github.io/datasets/}}.

\subsection{Source data}
All the raw, unprocessed data are exported and stored in the `\textit{SourceData}' folder. The subfolders within SourceData---Vicon, Noraxon, and ELONXI---correspond to data collected by different devices.
In the `\textit{Vicon}' subfolder, the kinematic and kinetic data are organized in a hierarchical structure of `Participant ID/Condition/*.c3d'.
In the `\textit{Noraxon}' subfolder, the sEMG data are stored in a hierarchical structure of `Participant ID/*.mat'.
In the `\textit{ELONXI}' subfolder, the AUS data are organized in a hierarchical structure of `Participant ID/Condition/Ambulation Mode/.txt'.
Similarly, more details about the `\textit{SourceData}' folder can be found at \href{https://k2muse.github.io/datasets/}{\textcolor{blue}{https://k2muse.github.io/datasets/}}.

\begin{table*}[htbp]
    \fontsize{8pt}{10pt}\selectfont
	\centering
	\caption{Datatype in the `OriginalData' structure, including unprocessed data without parsing or normalizing.}
	\label{tab:OriginalData}
	\resizebox{1\linewidth}{!}{
		\renewcommand{\arraystretch}{1.2} 
		\begin{tabular}{p{3cm}<{\raggedright} p{1.5cm}<{\raggedright} p{2cm}<{\raggedright} p{8.5cm}}
			\Xhline{1.2pt}  
			\textbf{Field within structure}	& \textbf{Units} 	& \textbf{Sampling rate} 	& \textbf{Contents} \\
			\Xhline{1.2pt}  
			{\ttfamily Markers}	&	(m)		&	100 Hz	&	Positions of the markers defined in Figure~\ref{fig:MarkerSet} in the global coordinate system.\\
			\cline{4-4}
			&&&	Array Format: (total frames $\times$ 3)\\
			\cline{4-4}
			&&& First Dimension: Frames in trial\\
			\cline{4-4}
			&&&	Second Dimension: x/y/z location in the global space\\
			\hline
			{\ttfamily Angles}	&	(deg)	&	100 Hz	&	Pelvic tilt, hip, knee, ankle and foot angles as defined by the Plug-in Gait Model.\\
			\cline{4-4}
			&&&	Array Format: (total frames $\times$ 3)\\
			\cline{4-4}
			&&&	First Dimension: Frames in trial\\
			\cline{4-4}
			&&&	Second Dimension: x/y/z rotation in local space\\
			\hline
			{\ttfamily Forces}	&	(N/kg)	&	100 Hz	& Forces acting on hip, knee, and ankle joints, expressed in the local coordinate frame of the distal segment.\\
			\cline{4-4}
			&&&	Array Format: (total frames $\times$ 3)\\
			\cline{4-4}
			&&&First Dimension: Frames in trial\\
			\cline{4-4}
			&&&	Second Dimension: x/y/z joint forces\\
			\hline
			{\ttfamily Moments}	&	(N.m/kg)	& 100 Hz	& Hip, knee, and ankle moments normalized by the participant's mass as defined by the Plug-in Gait model\\
			\cline{4-4}
			&&&	Array Format: (total frames $\times$ 3)\\
			\cline{4-4}
			&&&First Dimension: Frames in trial\\
			\cline{4-4}
			&&&	Second Dimension: x/y/z joint moments\\			
			\hline
			{\ttfamily Powers}	&	(W/kg)	&	100 Hz	& Powers at each joint normalized by the participant's mass, calculated as the multiplication of the joint moment and joint velocity.\\
			\cline{4-4}
			&&&	Array Format: (total frames $\times$ 3)\\
			\cline{4-4}
			&&&First Dimension: Frames in trial\\
			\cline{4-4}
			&&&	Second Dimension: x/y/z joint powers\\				
			\hline
			{\ttfamily Events}	& (s), (Frame)		&	N/A	& Gait events (heel strike and toe-off) detected by the force plates in each trial, presented in both time and frame number formats.\\
			\cline{4-4}
			&&&Array Format: (1 $\times$ HS/TO number)\\
			\cline{4-4}
			&&&	Second Dimension: Numbers of gait events.\\	
			\hline
			{\ttfamily USData}	&	mm	&	20 Hz	&	The extracted and merged AUS data, with frames shaped as (4, 1000), where rows 1 to 4 correspond to transducer channels 1 to 4.\\
			\cline{4-4}
			&&&	Array Format: (total frames $\times$ 1)\\
			\cline{4-4}
			&&&First Dimension: Frames in trial\\
			\hline
			{\ttfamily EMGData}	&	$\mu$$V$	&	2000 Hz	&	The extracted and merged sEMG data, where columns 1 to 13 correspond to sensor channels 1 to 13.\\
			\cline{4-4}
			&&&	Array Format: (total samples $\times$ 13)\\
			\cline{4-4}
			&&&First Dimension: Samples in trial\\
			\cline{4-4}
			&&&	Second Dimension: sEMG sensor channels\\	
			\Xhline{1.2pt}
		\end{tabular}
	}
\end{table*}

\begin{table*}[htbp]
	\fontsize{8pt}{10pt}\selectfont
	\centering
	\caption{Datatype in the `NormalizedData' structure, with the data parsed and time-normalized by strides.}
	\label{tab:NormalizedData}
	\resizebox{1\linewidth}{!}{
		\renewcommand{\arraystretch}{1.2} 
		\begin{tabular}{p{3cm}<{\raggedright} p{3cm}<{\raggedright} p{8cm}}
			\Xhline{1.2pt}  
			\textbf{Field within structure}	& \textbf{Sampling rate} 	& \textbf{Contents} \\
			\Xhline{1.2pt}  
			{\ttfamily Markers}	&	(m)	&	Array Format: (200 $\times$ 3 $\times$ stride)\\
			\cline{3-3}
			&& First Dimension: normalized gait cycle across the stride (200 points)\\
			\cline{3-3}
			&&	Second Dimension: x/y/z location in the global space\\
			\cline{3-3}
			&&	Third Dimension: stride number\\			
			\hline
			{\ttfamily Angles}	&	(deg)	&	Array Format: (200 $\times$ 3 $\times$ stride)\\
			\cline{3-3}
			&&	First Dimension: normalized gait cycle across the stride (200 points)\\
			\cline{3-3}
			&&	Second Dimension: x/y/z rotation in local space\\
			\cline{3-3}
			&&	Third Dimension: stride number\\
			\hline
			{\ttfamily Forces}	&	(N/kg)	&	Array Format: (200 $\times$ 3 $\times$ stride)\\
			\cline{3-3}
			&&	First Dimension: normalized gait cycle across the stride (200 points)\\
			\cline{3-3}
			&&	Second Dimension: x/y/z joint forces\\
			\cline{3-3}
			&& Third Dimension: stride number\\
			\hline
			{\ttfamily Moments}	&	(N.m/kg)	&	Array Format: (200 $\times$ 3 $\times$ stride)\\
			\cline{3-3}
			&&	First Dimension: normalized gait cycle across the stride (200 points)\\
			\cline{3-3}
			&&	Second Dimension: x/y/z joint moments\\
			\cline{3-3}
			&& Third Dimension: stride number\\
			\hline
			{\ttfamily Powers}	&	(W/kg)	&	Array Format: (200 $\times$ 3 $\times$ stride)\\
			\cline{3-3}
			&&	First Dimension: normalized gait cycle across the stride (200 points)\\
			\cline{3-3}
			&&	Second Dimension: x/y/z joint powers\\
			\cline{3-3}
			&&	Third Dimension: stride number\\
			\hline
			{\ttfamily Events}	&	(Frame)	&	Array Format: (1 $\times$ HS number)\\
			\cline{3-3}
			&&	Second Dimension: stride number\\
			\hline
			{\ttfamily USData}	&	mm	&	Array Format: (stride $\times$ 1)\\
			\cline{3-3}
			&&First Dimension: stride number\\
			\hline
			{\ttfamily EMGData}	&	$\mu$$V$	&	Array Format: (stride $\times$ 1)\\
			\cline{3-3}
			&&First Dimension: stride number\\
			\Xhline{1.2pt}
		\end{tabular}
	}
\end{table*}

\section{Analysis and validation}
\label{AnalysisandValidation}

\subsection{Reliability of the Plug-in Gait protocol}
The Plug-in Gait marker set employed in this study exhibits excellent intra-protocol repeatability and is widely utilized in gait analysis. 
The primary source of variability arises from differences in marker placement, which remains the key factor contributing to motion capture discrepancies \citep{MarkerVariability}.
The marker placement procedure, which is based on anatomical landmarks, was carefully designed to minimize variability. To ensure reliability and consistency, two assessors were responsible for marker placement across the entire dataset.
Since two types of tape were used for fixation, marker detachment occurred only in rare cases. If a marker fell off, it was promptly reattached at the exact same position, guided by the imprint left on the skin.
The validated Vicon Plug-in Gait model was utilized to ensure reliable marker placement \citep{PiG_1, PiG_2}. The standard and widely adopted Plug-in Gait Dynamic pipeline in the Vicon Nexus software was employed for motion data processing, ensuring high reliability \citep{PiG_Reliability}.

\subsection{Motion capture and sensor data}
Before each experimental session, the motion capture system was calibrated according to the manufacturer's standard procedure, which involved calibrating the capture volume and setting the volume origin. The system was recalibrated whenever any camera was unintentionally disturbed by external factors.
The force plates in the Bertec treadmill were subsequently reset, including leveling the treadmill and setting the level to zero in the Nexus software. The force plates were zeroed in the Nexus software each time the treadmill incline was changed.
The retro-reflective markers, sEMG sensors and AUS transducers were positioned through palpation of bony landmarks and muscle tissue, according to the guidelines of \citet{SENIAM} and \citet{MarkerPosition}.
The sEMG and AUS signals were inspected at the beginning of each trial and monitored throughout the experiment, ensuring consistent quality.

\subsection{Synchronization}
\begin{figure*}
	\centering
	{\includegraphics[width=2\columnwidth]{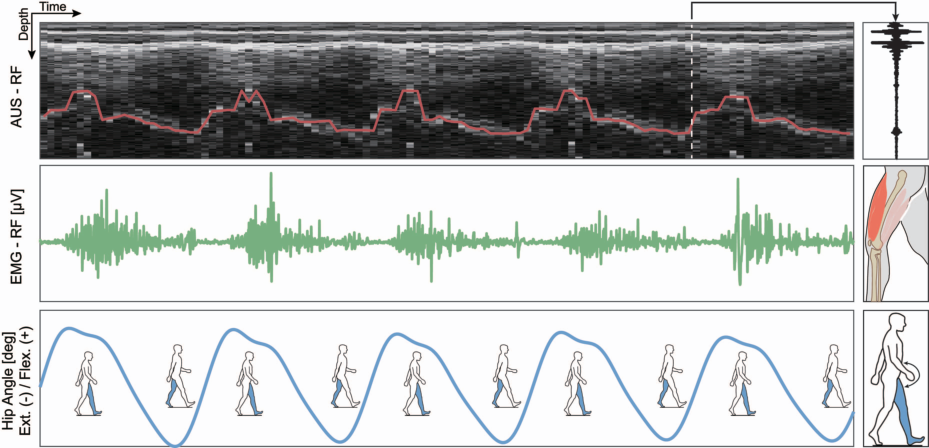}}
	\captionsetup{justification=justified, singlelinecheck=false}
	\caption{During level-ground walking at 1.0 m/s, the representative AUS and sEMG data of the left rectus femoris (RF), along with the left hip joint angle data, were recorded. The insets on the right side of the AUS data present representative raw AUS data captured at specific time frames.}
	\label{fig:MuscleTracking}
\end{figure*}

The data collected by different devices during steady-state walking at 1.0 m/s on level ground by a participant, including the AUS and sEMG data of the left leg's rectus femoris (RF) and the left hip joint angle, are shown in Figure~\ref{fig:MuscleTracking}.
For the AUS data, a brightness-based method was utilized to track variations in muscle thickness \citep{AUS_Torque}.
Following the heel strike of the left foot, the hip joint angle decreases from its peak value. During this phase, the left leg supports the body, leading to RF contraction, an increase in muscle thickness, and an increase in the sEMG signal amplitude.  
As the left hip joint angle reaches its minimum value, the center of gravity shifts toward the right leg. At this point, the muscle thickness of the left RF reaches its maximum, and the sEMG signal amplitude begins to decrease.  
Between the current toe-off event and the next heel strike event of the left foot, the hip joint angle increases because of body inertia. During this period, the muscle does not exert significant force, resulting in a reduction in both muscle thickness and sEMG signal amplitude until the next heel strike event occurs.  
Therefore, according to the above analysis, the simultaneous changes in the joint angle, muscle thickness, and sEMG signal intensity can be used to validate the synchronization among the different devices.

\begin{figure}
	\centering
	{\includegraphics[width=1\columnwidth]{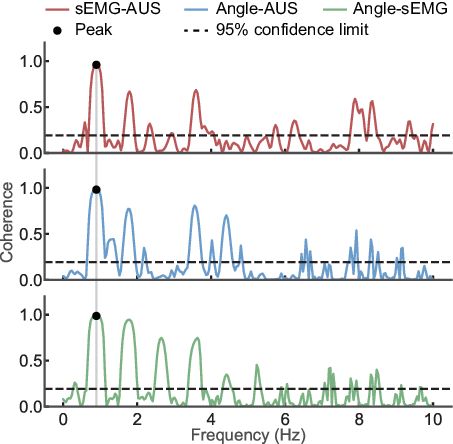}}
	\captionsetup{justification=justified, singlelinecheck=false}
	\caption{Coherence spectra between multimodal signal pairs from the left rectus femoris, showing frequency-dependent coupling across modalities. The black horizontal dashed line denotes the 95\% confidence limit. The gray vertical solid line indicates the peak frequency of the coherence spectrum.}
	\label{fig:TimeFreqCoherence}
\end{figure}

To further quantify the dynamic coupling among sEMG, AUS, and joint angle, we employed the magnitude squared coherence (MSC) analysis \citep{mima2001coherence, grosse2004patterns}. 
The MSC at frequency $f$ is defined as the normalized ratio of the cross-power spectral density to the auto-power spectral density as follows:
\begin{equation}
	\operatorname{MSC}_{xy}(f) = \frac{|P_{xy}(f)|^2}{P_{xx}(f)\,P_{yy}(f)}, \quad 0 \le \operatorname{MSC}_{xy}(f) \le 1
\end{equation}
where $P_{xy}(f)$ denotes the cross-power spectral density between signals $x$ and $y$, and $P_{xx}(f)$ and $P_{yy}(f)$ denote the auto-power spectral densities of $x$ and $y$, respectively. 
Spectral densities were estimated using Welch's averaged, segmented periodogram method. To assess statistical significance, the 95\% confidence limit was set to
\begin{equation}
	C_{95}=1-\alpha^{\frac{1}{K-1}},\quad \alpha=0.05
\end{equation}
where $K$ denotes the effective number of segments used in Welch's method for spectral estimation.
Figure~\ref{fig:TimeFreqCoherence} shows the coherence spectra of the left RF during level-ground walking at 1.0 m/s after the multimodal signals were temporally aligned as shown in Figure~\ref{fig:MuscleTracking}. 
Three modality pairs were considered, including sEMG and AUS, angle and AUS, as well as angle and sEMG. 
The black horizontal dashed lines denote the 95\% confidence limits, and the circular markers indicate the coherence peak for each curve.
All modality pairs exhibited significant coherence in the low-frequency range, and the dominant peak occurred at a consistent frequency of approximately 0.90 Hz. The corresponding peak coherence values were 0.961 for the sEMG and AUS pair, 0.981 for the angle and AUS pair, and 0.987 for the angle and sEMG pair, all of which were well above the 95\% confidence limit. 
These results suggest a strong association between the electrical muscle activity captured by sEMG signals and the muscle thickness variations measured by AUS. Furthermore, these results indicate that both modalities track the periodic joint-angle modulation during gait. Consistent cross-modal synchronization around the fundamental gait rhythm was observed.
From a frequency-domain perspective, the high coherence supports the consistency of contraction-related dynamics across sensing modalities and highlights a shared dominant rhythm. 
These findings provide a reliable basis and quantitative spectral metrics for future studies on muscle force generation and multimodal modeling.

\begin{figure}
	\centering
	{\includegraphics[width=1\columnwidth]{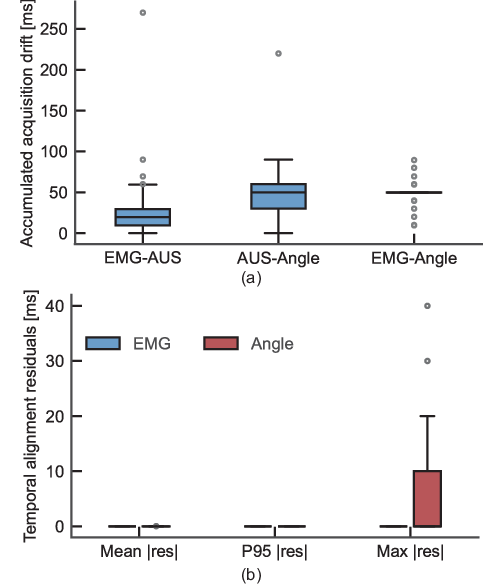}}
	\captionsetup{justification=justified, singlelinecheck=false}
	\caption{Synchronization error analysis of the multimodal acquisition system. (a) Total-duration drift between devices, which characterizes the long-term stability of hardware synchronization among the sEMG, AUS, and angle streams. (b) Timestamp alignment residuals, which quantify the instantaneous timing error when each AUS frame timestamp is mapped to the nearest sEMG or angle sample. The mean, 95th percentile (P95), and maximum (Max) residuals are reported.}
	\label{fig:SyncCheck}
\end{figure}

To quantitatively assess the synchronization performance of the multimodal hardware systems, we conducted a timestamp-based cross-validation analysis. 
We calculated the duration consistency and instantaneous alignment residuals across different modalities, as presented in Figure~\ref{fig:SyncCheck}.
The total duration drift among the sEMG, AUS, and motion capture systems within each trial, which characterize the synchronization stability over the course of data acquisition, is shown in Figure~\ref{fig:SyncCheck}(a).
The results show that, for the vast majority of trials, the accumulated drift remained on the order of tens of milliseconds, with a stable distribution and only a small number of outliers. Given that gait kinematics are dominated by low-frequency components, such small and approximately linear temporal deviations are unlikely to materially affect multimodal signal fusion.
The alignment residuals computed using the timestamp-based nearest-neighbor matching approach are shown in Figure~\ref{fig:SyncCheck}(b). Specifically, each valid AUS frame timestamp was mapped to the nearest EMG or Vicon sample, and the resulting temporal alignment residuals were aggregated across frames. 
The mapping residuals between the AUS and EMG signals were negligible. 
The residuals between the AUS signals and joint angles were also small, with a maximum of approximately 40 ms. 
Overall, most synchronization errors were well below the AUS sampling interval. These results indicate that the square-wave trigger generated by the Vicon Lock Lab effectively constrained the time bases across devices and enabled multimodal alignment with millisecond-level accuracy.

\subsection{Kinematic and kinetic data}

\begin{figure*}
	\centering
	{\includegraphics[width=1\linewidth]{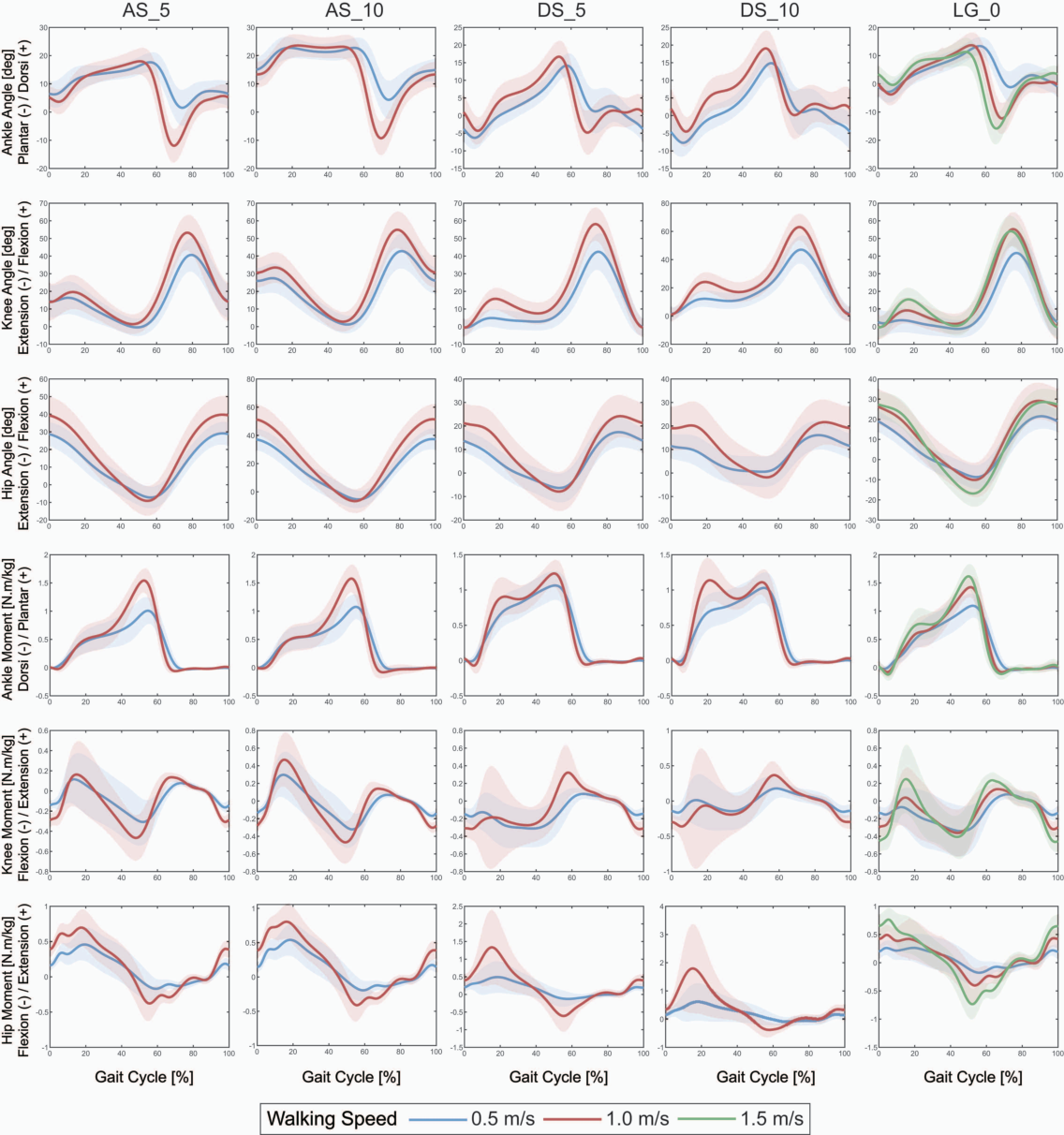}}
	\captionsetup{justification=justified, singlelinecheck=false}
	\caption{Joint angles and moments during ideal condition experiments. Two consecutive heel strikes correspond to 0\% and 100\% of the gait cycle. The ambulation mode is annotated at the top. The solid lines indicate the average trajectory across all participants. The shaded regions correspond to the standard deviation.}
	\label{fig:IdealKinematicKinetic}
\end{figure*}

\begin{figure*}
	\centering
	{\includegraphics[width=1\linewidth]{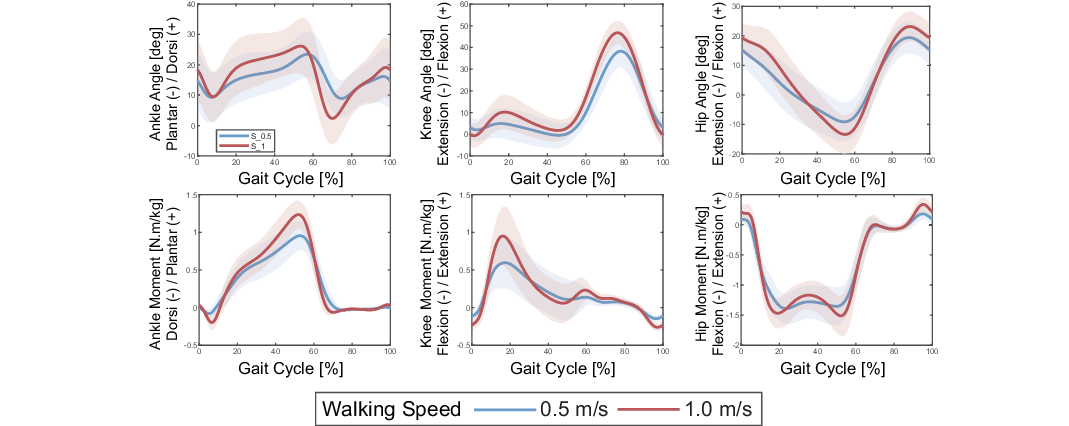}}
	\captionsetup{justification=justified, singlelinecheck=false}
	\caption{Joint angles and moments of older participants during level-ground walking at different speeds in the ideal-condition experiments. Two consecutive heel strikes correspond to 0\% and 100\% of the gait cycle. The ambulation mode is annotated at the top. The solid lines indicate the average trajectory across all participants. The shaded regions correspond to the standard deviation.}
	\label{fig:IdealKinematicKineticOlder}
\end{figure*}

Although the participants were able-bodied, perfect bilateral symmetry cannot be assumed in the human gait \citep{maqbool2024gait}. Therefore, for consistency and to avoid redundancy in presentation, we report the kinematic and kinetic profiles of the right leg in the main text, while bilateral measurements are available in the dataset for users who wish to analyze inter-limb asymmetry.
The joint angles and moments of healthy participants in the sagittal plane under ideal conditions are depicted in Figure~\ref{fig:IdealKinematicKinetic}. Positive joint angle values represent ankle dorsiflexion, knee flexion, and hip flexion, whereas positive joint moment values denote ankle plantar flexion, knee extension, and hip extension.
The sagittal-plane joint angles and moments of the older participants are shown in Figure~\ref{fig:IdealKinematicKineticOlder}. Compared with the healthy group, older participants exhibited distinct kinematic patterns under ideal walking conditions. 
Under the same experimental paradigm, the cross-correlation coefficient (XCOR) \citep{KK_HDEMG} and Pearson correlation coefficient (PCC) \citep{KK_EMG2} of the joint angles and joint moments between the healthy and older cohorts are summarized in Table~\ref{tab:HealthyVsElderly}.
Overall, compared with the healthy participants, the sagittal-plane ranges of motion at the hip and knee were reduced in the older group.
Joint moments also differed between the two groups.
Although the ankle moment profiles were broadly comparable, more pronounced differences were observed between the hip and knee in the older group.
Additionally, the knee moment exhibited a relatively large peak at approximately 20\% of the gait cycle.
In addition, compared with the healthy group, the hip moment distribution across flexion and extension was less symmetric in the older group, with a clear bias toward the flexion direction.

\begin{table}[htb]
	\fontsize{8pt}{10pt}\selectfont
	\centering
	\captionsetup{justification=justified, singlelinecheck=false}
	\caption{Summary of the XCOR and PCC values between the healthy and elderly groups for the right sagittal plane joint angles and moments under level-ground walking.}
	\label{tab:HealthyVsElderly}
	\renewcommand{\arraystretch}{1.15}
	\resizebox{\columnwidth}{!}{
		\begin{tabular}{c c c c c c c c}
			\toprule[1.2pt]
			\multirow{2}{*}{\textbf{Data}} & \multirow{2}{*}{\textbf{Metric}} &
			\multicolumn{3}{c}{\textbf{0.5 m/s}} & \multicolumn{3}{c}{\textbf{1.0 m/s}} \\
			\cmidrule(lr){3-5}\cmidrule(lr){6-8}
			& & \textbf{Ankle} & \textbf{Knee} & \textbf{Hip} & \textbf{Ankle} & \textbf{Knee} & \textbf{Hip} \\
			\midrule[1.2pt]
			
			\multirow{2}{*}{\textbf{Angle}}
			& XCOR & 0.8270 & 0.9979 & 0.9941 & 0.6268 & 0.9983 & 0.9763 \\
			\cline{2-8}
			& PCC  & 0.9392 & 0.9985 & 0.9963 & 0.9704 & 0.9957 & 0.9976 \\
			\midrule
			
			\multirow{2}{*}{\textbf{Moment}}
			& XCOR & 0.9986 & 0.0596 & 0.2694 & 0.9973 & 0.2075 & 0.3367 \\
			\cline{2-8}
			& PCC  & 0.9967 & -0.2141 & -0.0883 & 0.9899 & 0.3359 & 0.1721 \\
			\bottomrule[1.2pt]
	\end{tabular}}
\end{table}

The repeatability of the data reflects the consistency in the participants' movements, which directly affects the performance of the intention recognition approach and the reliable operation of robotic systems \citep{LiGuangLin}. To ensure this repeatability, the joint angles across participants were validated through the following procedure.
Let $p$ denote the participant and $m$ denote the ambulation mode. For mode $m$, participant $p$ has $N_{p,m}$ gait cycles.
For each joint $j\in\{\mathrm{hip},\mathrm{knee},\mathrm{ankle}\}$, let $\theta^{(j)}_{p,m,i}(t)$ be the joint-angle trajectory of the $i$-th gait cycle ($i=1,\dots,N_{p,m}$) sampled over a normalized gait cycle $t=1,\dots,T$.

First, for each participant and ambulation mode, the mean joint-angle trajectory across gait cycles was computed as follows:
\begin{equation}
	\bar{\theta}^{(j)}_{p,m}(t)=\frac{1}{N_{p,m}}\sum_{i=1}^{N_{p,m}}\theta^{(j)}_{p,m,i}(t)
\end{equation}

Next, the coefficient of determination between each gait cycle trajectory and the corresponding mean trajectory were computed as follows:
\begin{equation}
\begin{aligned}
	R^{2\,(j)}_{p,m,i}
	&=1-\frac{\sum_{t=1}^{T}\left(\theta^{(j)}_{p,m,i}(t)-\bar{\theta}^{(j)}_{p,m}(t)\right)^2}
	{\sum_{t=1}^{T}\left(\theta^{(j)}_{p,m,i}(t)-\mu^{(j)}_{p,m,i}\right)^2}\\
	\mu^{(j)}_{p,m,i}
	&=\frac{1}{T}\sum_{t=1}^{T}\theta^{(j)}_{p,m,i}(t)
\end{aligned}
\end{equation}
and these values were averaged across gait cycles to obtain a joint-wise score as follows:
\begin{equation}
	\overline{R^2}^{\,(j)}_{p,m}=\frac{1}{N_{p,m}}\sum_{i=1}^{N_{p,m}}R^{2\,(j)}_{p,m,i}
\end{equation}

A range-of-motion weight was then assigned to each joint according to its relative motion amplitude in mode $m$:
\begin{equation}
\begin{aligned}
	\Delta^{(j)}_{p,m}
	&=\max_{t}\bar{\theta}^{(j)}_{p,m}(t)-\min_{t}\bar{\theta}^{(j)}_{p,m}(t)\\
	w^{(j)}_{p,m}
	&=\frac{\Delta^{(j)}_{p,m}}{\sum_{k\in\{\mathrm{hip},\mathrm{knee},\mathrm{ankle}\}}\Delta^{(k)}_{p,m}}
\end{aligned}
\end{equation}

Finally, the participant-level score for each ambulation mode was obtained by weighting the joint-wise $\overline{R^2}$ values by the corresponding range-of-motion weights:
\begin{equation}
	\overline{R^2}_{p,m}=\sum_{j\in\{\mathrm{hip},\mathrm{knee},\mathrm{ankle}\}}w^{(j)}_{p,m}\,\overline{R^2}^{\,(j)}_{p,m}
\end{equation}

\begin{figure}
	\centering
	{\includegraphics[width=0.95\columnwidth]{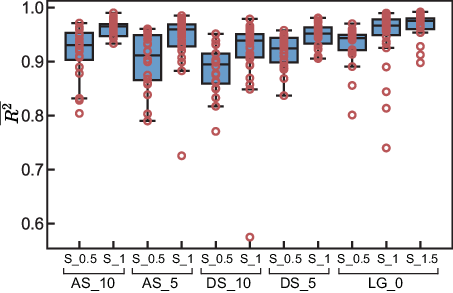}}
	\captionsetup{justification=justified, singlelinecheck=false}
	\caption{The evaluation of data repeatability. The circles represent the $\overline{R^2}$ between the average angle and joint angles for each participant across different ambulation modes. The box represents the distribution of $\overline{R^2}$ for each movement.}
	\label{fig:Repeatability}
\end{figure}

Figure~\ref{fig:Repeatability} presents the distribution of $\overline{R^2}$ values for the angles across the five walking modes, with most values exceeding 0.8, demonstrating good repeatability.

\subsection{Non-ideal condition analysis} \label{NonidealAnalysis}

As described in Section~\ref{Experimental protocol}, the K2MUSE dataset includes several experiments conducted under common non-ideal conditions, including muscle fatigue conditions, electrode shifts, and inter-day differences. Such factors are prevalent in clinical and home-use rehabilitation robotics because repeated donning and doffing, contact pressure changes, fatigue accumulation, and day-to-day variability in placement and skin--electrode impedance can introduce distribution shifts in sEMG and related signals.
These shifts can decrease intent-decoding accuracy and control stability when recalibration is not performed \citep{li2020electrode, li2025optimizing, EMGFeature}.
By providing synchronized sEMG, AUS, kinematic, and kinetic data and by explicitly including these representative non-ideal scenarios, the K2MUSE dataset enables benchmarking of sensing variability and supports the development and evaluation of robust multimodal models and adaptation strategies for out-of-lab deployment.
Accordingly, in this section, the effects of non-ideal conditions on signal acquisition are investigated, using sEMG as a representative example.

\subsubsection{Muscle fatigue.} 
Muscle fatigue can lead to feature drift in human--machine interfaces, as sEMG-based features are sensitive to fatigue states \citep{Fatigue_Status}.
The median frequency (MDF) and mean frequency (MF) of sEMG signals are widely recognized as reliable indicators for assessing muscle fatigue, as both have been shown to decrease with increasing fatigue level \citep{MDF_MF}.

\begin{figure*}
	\centering
	{\includegraphics[width=1\linewidth]{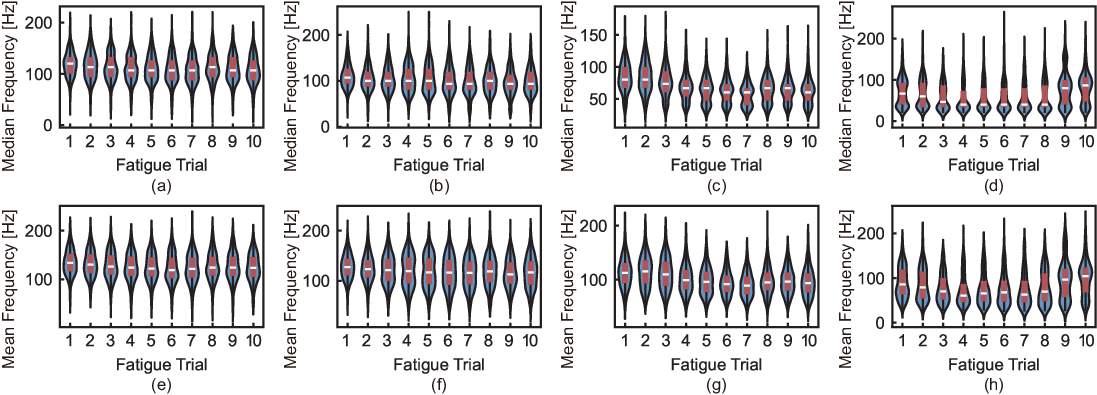}}
	\captionsetup{justification=justified, singlelinecheck=false}
	\caption{Violin plots illustrate the distributions of the median frequency and mean frequency for different muscles during the progressive fatigue trials induced by exercise. (a)--(d) show the median frequencies of the TA, LG, RF, and BF, respectively. (e)--(h) show the mean frequencies of the TA, LG, RF, and BF, respectively.}
	\label{fig:FatigueMetric_LG}
\end{figure*}

The results of the progressive fatigue trials are shown in Figure~\ref{fig:FatigueMetric_LG}. As the trials progressed, the median frequency and mean frequency of the TA and LG remained relatively stable, with only a slight decreasing trend. In contrast, the median frequency and mean frequency of the RF clearly decreased during the early stage of the protocol, and these metrics remained at lower levels in the later trials. This pattern is consistent with a fatigue-related shift of the power spectrum toward lower frequencies. The frequency metrics of the BF showed more pronounced non-monotonic behavior, with greater decreases in the middle stage and partial rebounds in the final stage. This trend agrees with the observations reported in \citet{MDF_MF}.

\begin{figure*}
	\centering
	{\includegraphics[width=1\linewidth]{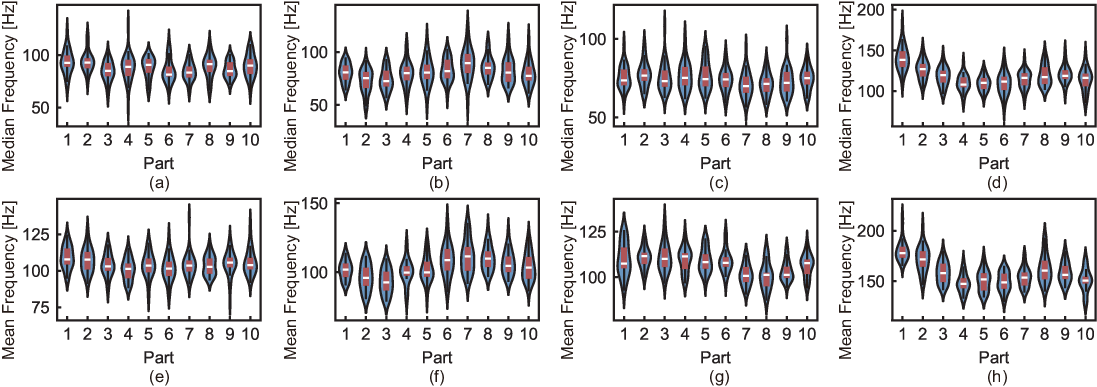}}
	\captionsetup{justification=justified, singlelinecheck=false}
	\caption{Violin plots illustrate the distributions of the median frequency and mean frequency for different muscles during the sustained fatigue trial induced by continuous uphill walking. (a)--(d) show the median frequencies of the TA, LG, RF, and BF, respectively. (e)--(h) show the mean frequencies of the TA, LG, RF, and BF, respectively.}
	\label{fig:FatigueMetric_AS}
\end{figure*}

The results of the sustained fatigue trial are shown in Figure~\ref{fig:FatigueMetric_AS}. The continuous 6 min uphill walking data were evenly divided into ten time segments (Part 1 to Part 10). Within each segment, we first computed the median frequency and mean frequency for each gait cycle and then aggregated these values to form the corresponding distributions. This procedure was used to characterize how fatigue evolved over time. For the TA and LG, both the median frequency and mean frequency exhibited limited fluctuations. Only mild decreases or segment-specific rebounds were observed, suggesting that the spectral shift toward lower frequencies was relatively small during continuous uphill walking. In contrast, more pronounced changes were observed for the RF and BF. With respect to the RF, the frequency metrics clearly decreased in the first half of the trial and remained relatively low in the later segments. With respect to the BF, the metrics substantially decreased in the early segments, followed by a partial rebound during the middle and late stages.

\begin{figure}
	\centering
	{\includegraphics[width=0.9\columnwidth]{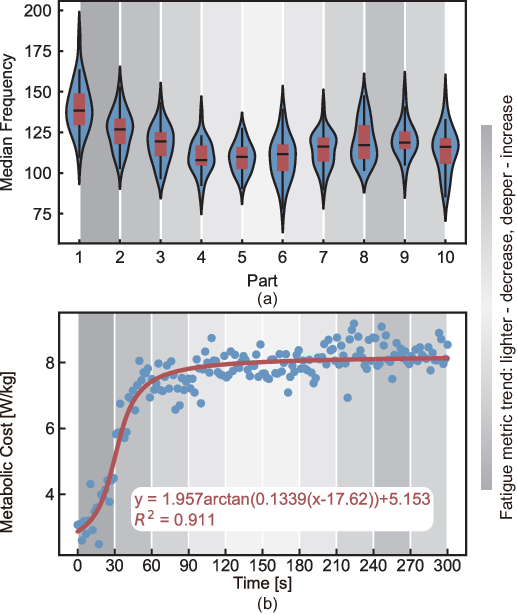}}
	\captionsetup{justification=justified, singlelinecheck=false}
	\caption{Trends of (a) the BF median frequency and (b) participants' metabolic cost during the sustained fatigue trial induced by continuous uphill walking.}
	\label{fig:FatigueMetric_AS_CMP}
\end{figure}

As shown in Figure~\ref{fig:FatigueMetric_AS_CMP}, we further investigated the fatigue phenomenon caused by continuous uphill walking from the perspectives of muscle fatigue and overall metabolic cost. The distribution of the BF median frequency across the 6 min trial after the data were evenly divided into ten time segments (Part 1 to Part 10) is shown in Figure~\ref{fig:FatigueMetric_AS_CMP}(a). The temporal evolution of metabolic cost is shown in Figure~\ref{fig:FatigueMetric_AS_CMP}(b). The metabolic cost increased rapidly at the beginning of the trial but then gradually approached a steady level. This evolution was well captured by an arctangent model ($R^2 = 0.911$), indicating a progressive stabilization of the metabolic load during continuous uphill walking. Taken together, the data in Figure~\ref{fig:FatigueMetric_AS_CMP}(a) and (b) suggest that as the trial progressed, participants exhibited clearer signs of muscle fatigue, while the whole-body metabolic cost increased and subsequently stabilized. These coupled changes reflect the relationship between muscle state and systemic energy demand during the fatigue process. They also support the physiological validity of the fatigue-related recordings in the K2MUSE dataset.

\subsubsection{Electrode Shifts.}
To simulate electrode shifts, four offset positions were established in different directions around the initial position. 
The mean absolute value (MAV) was selected as the metric for measuring the sEMG amplitudes of the four muscles in the left leg \citep{SeNic}, which was calculated as follows:
\begin{equation}
	MAV = \frac{1}{N_t} \sum_{i=1}^{N_t} |x_i|
\end{equation}
where $N_t$ denotes the analysis window length, $N_c$ denotes the increment length between adjacent windows, and $x_i$ denotes the $i$-th sEMG sample within the window \citep{khezri2010neuro}.

\begin{figure}
	\centering
	{\includegraphics[width=0.9\columnwidth]{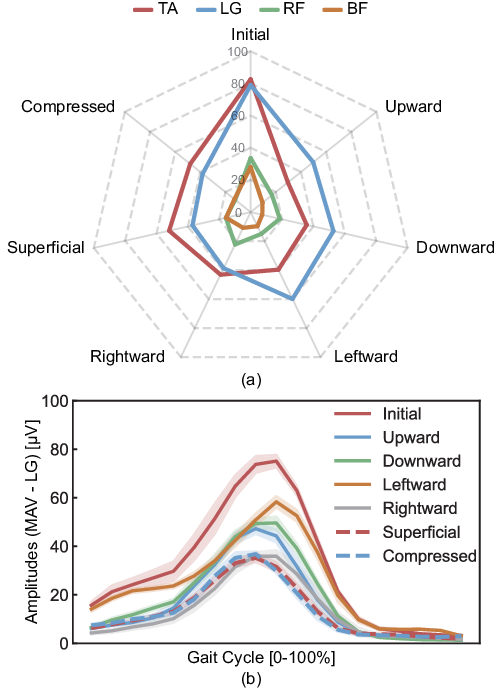}}
	\captionsetup{justification=justified, singlelinecheck=false}
	\caption{For the different shift conditions: (a) the mean of the maximum MAV across all gait cycles for sEMG signals recorded from four muscles of the left leg; (b) MAV variation of the LG in the left leg throughout the gait cycle. The solid lines represent the mean MAVs across all cycles for different two-dimensional shift positions, whereas the dashed lines represent the mean MAVs across all cycles for different contact pressure levels. The shaded regions indicate one standard deviation. Note that the data collected under different contact pressure levels and the data collected under different two-dimensional shift positions were acquired on different days.}
	\label{fig:ShfitRadarCurve}
\end{figure}

As shown in Figure~\ref{fig:ShfitRadarCurve}~(a), for all five repeated trials of a participant at different shift positions, the maximum muscle action potentials were determined by averaging the maximum MAVs across all the gait cycles.
For the four muscles, the maximum sEMG amplitudes were recorded at the initial position. The amplitudes for the shank muscles (TA and LG) were smallest at the upward and rightward shift positions, whereas the amplitudes for the thigh muscles (RF and BF) were smallest at the leftward and downward offset positions.
Additionally, Figure~\ref{fig:ShfitRadarCurve}~(b) illustrates the amplitude variation of the LG throughout the gait cycles, with the curve for the initial position consistently above those for the other shift positions.

In the second component, depth-direction variations were induced by modulating the electrode--skin contact pressure under two conditions, namely, superficial and compressed conditions (dashed curves in Figure~\ref{fig:ShfitRadarCurve}). When these two pressure levels were compared within the same session, the MAV patterns were largely consistent. Both the maximum MAV summarized in Figure~\ref{fig:ShfitRadarCurve} (a) and the LG MAV trajectories over the gait cycle in Figure~\ref{fig:ShfitRadarCurve} (b) showed only minor differences between the superficial and compressed conditions, and their distributions overlapped substantially. Because the pressure modulation trials and the in-plane shift trials were conducted on different days, we restricted the analysis to within-component comparisons to reduce the influence of day-to-day variability.

\subsubsection{Inter-day difference.}
\begin{figure}
	\centering
	{\includegraphics[width=0.9\columnwidth]{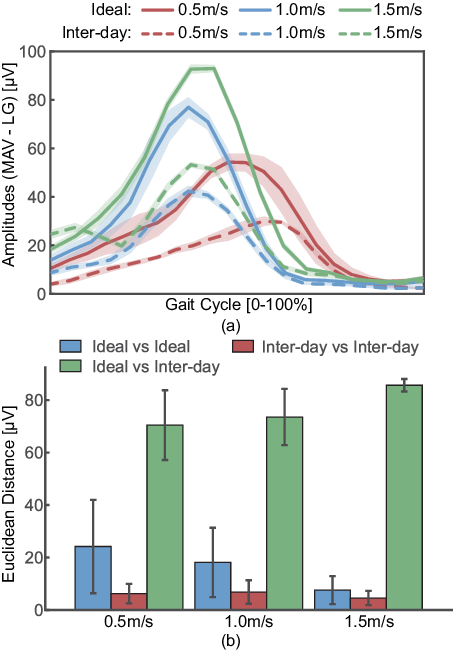}}
	\captionsetup{justification=justified, singlelinecheck=false}
	\caption{For ideal conditions and inter-day differences: (a) The variation in the MAVs of the left leg LG throughout the gait cycle. The solid lines represent the average values across all gait cycles from five walking trials, whereas the shaded regions indicate the standard deviation. (b) Comparison of the Euclidean distance of the MAVs for the left leg LG under two conditions: intra-condition trial comparisons and inter-condition trial comparisons. The bar height represents the mean value, and the error bars indicate the standard deviation.}
	\label{fig:InterDayCurveBar}
\end{figure}

The inter-day difference experiments were performed on a separate day from the ideal condition experiments. Consequently, a comparative analysis of the sEMG data under these two conditions was performed. We examined the MAVs of the sEMG signals across all the gait cycles from five walking trials of a participant at walking speeds of 0.5 m/s, 1.0 m/s, and 1.5 m/s.
As shown in Figure~\ref{fig:InterDayCurveBar}~(a), for the three walking speeds, the MAV variations throughout the gait cycle exhibit similar trends under both conditions. However, significant differences in amplitude were observed, indicating that while the participant's walking pattern remained consistent across conditions, the characteristics of the sEMG signal changed.
Specifically, the Euclidean distance shown in Figure~\ref{fig:InterDayCurveBar}~(b) provides a quantitative assessment of these characteristics. The intra-condition distances for both the ideal condition and inter-day difference experiments remain relatively small, indicating consistency in the sEMG data characteristics across the five trials within each condition. However, the significantly greater Euclidean distance between the ideal condition and inter-day difference condition suggests that variations in the participant's physiological state on different days result in notable changes in the sEMG signal characteristics.

\subsection{Comparison with public datasets}

To evaluate the validity of the data from walking on different terrains, we compared ankle, knee, and hip angles from the young adult cohort with those from two publicly available datasets.
For the dataset from \citet{SciDataGregg}, the comparison includes walking data on various ramps, with all the walking speeds set to 1.0 m/s. The participants in this dataset were healthy, with an average age of 30.4 years.
In the dataset from \citet{CyclicTasks}, owing to differences in experimental paradigms, walking on ramps at 1.0 m/s was compared, whereas walking on level ground at 1.2 m/s was compared. The population in this dataset had an average age of 21.8 years, which is similar to that of our study.

We calculated the average values of our data, along with the cross-correlation coefficients (XCORs) and Pearson correlation coefficients (PCCs) with the selected comparison datasets, as shown in Table~\ref{tab:Comparison}.
The results demonstrate that the K2MUSE dataset is highly correlated with the datasets used in previous studies. 
For the data from \citet{SciDataGregg}, the XCOR and PCC values for the joint angles range from 0.88 to 0.99 and from 0.93 to 0.99, respectively. 
Similarly, for the data from \citet{CyclicTasks}, the XCOR and PCC values for the joint angles range from 0.81 to 0.99 and from 0.90 to 0.99, respectively.

In addition, we examined whether the treadmill-based recordings exhibit kinematic patterns that are consistent with natural overground walking. Specifically, the level-ground treadmill condition at 1.0 m/s in the K2MUSE dataset was compared with similar conditions in several publicly available overground walking datasets. As shown in Table~\ref{tab:Comparison_LG}, the ankle, knee, and hip joint angles demonstrated consistently high similarity (XCOR: 0.88--0.99; PCC: 0.80--0.98). 
These findings provide evidence that the treadmill-based kinematic data are broadly comparable to those reported for overground gait, supporting the use of the K2MUSE dataset as a baseline dataset for subsequent data-driven gait analyses.

For walking tasks comparable to those in other datasets, we compared the trends in kinematic and kinetic variations.
Compared with the dataset from \citet{Ramp_Stair}, the kinematic and kinetic variations associated with walking on level ground in our dataset are similar. 
In kinematic analysis, the hip joint angle follows a nearly sinusoidal pattern, ranging from approximately 20$^\circ$ of flexion to 20$^\circ$ of extension.
Similarly, the range of motion of the ankle joint motion spans from 15$^\circ$ of dorsiflexion to 15$^\circ$ of plantar flexion, and the knee joint kinematics range from 0$^\circ$ to 60$^\circ$.
Additionally, compared with \citet{KK_HDEMG}, similar kinematic and kinetic trends are observed during walking on 5$^\circ$ ascending and descending ramps in our dataset.

\begin{table*}[htb]
	\fontsize{8pt}{10pt}\selectfont
	\centering
	\captionsetup{justification=justified, singlelinecheck=false}
	\caption{Summary of the XCOR and PCC results for treadmill walking data from the K2MUSE dataset and the comparison dataset. Abbreviations: level ground (LG), descending ramp (DS), and ascending ramp (AS). The numeric suffix indicates the ramp inclination in degrees.}
	\label{tab:Comparison}
	\resizebox{1\linewidth}{!}{
		\renewcommand{\arraystretch}{1.2} 
		\begin{tabular}{c l ccc ccc ccc ccc ccc}
			\toprule[1.2pt]
			\multirow{2}{*}{\textbf{Dataset}} & \multirow{2}{*}{\textbf{Metric}}	&	\multicolumn{3}{c}{\textbf{LG\_0}}	&	\multicolumn{3}{c}{\textbf{AS\_5}}	&	\multicolumn{3}{c}{\textbf{AS\_1}0}	&	\multicolumn{3}{c}{\textbf{DS\_5}}	&	\multicolumn{3}{c}{\textbf{DS\_10}}\\
			\cmidrule(lr){3-5} \cmidrule(lr){6-8} \cmidrule(lr){9-11} \cmidrule(lr){12-14} \cmidrule(lr){15-17}
			&			&	\textbf{Ankle}	&	\textbf{Knee}	&	\textbf{Hip}	&	\textbf{Ankle}	&	\textbf{Knee}	&	\textbf{Hip}	&	\textbf{Ankle}	&	\textbf{Knee}	&	\textbf{Hip}	&	\textbf{Ankle}	&	\textbf{Knee}	&	\textbf{Hip}	&	\textbf{Ankle}	&	\textbf{Knee}	&	\textbf{Hip}\\
			\midrule[1.2pt]
			
			\multirow{2}{*}{\citet{SciDataGregg}}	&	XCOR	&	0.93	&	0.98	&	0.99	&	0.93	&	0.97	&	0.99	&	0.98	&	0.97	&	0.98	&	0.90	&	0.97	&	0.96	&	0.88	&	0.97	&	0.90	\\ 
			\cline{2-17}
			&	PCC		&	0.93	&	0.96	&	0.99	&	0.95	&	0.96	&	0.98	&	0.96	&	0.97	&	0.99	&	0.96	&	0.97	&	0.99	&	0.93	&	0.96	&	0.98	\\
			\midrule
			\multirow{2}{*}{\citet{CyclicTasks}}		&	XCOR	&	0.82	&	0.99	&	0.99	&	0.82	&	0.99	&	0.99	&	0.91	&	0.99	&	0.99	&	0.81	&	0.99	&	0.98	&	0.81	&	0.99	&	0.98	\\
			\cline{2-17}
			&	PCC		&	0.93	&	0.94	&	0.98	&	0.92	&	0.92	&	0.98	&	0.95	&	0.94	&	0.99	&	0.91	&	0.93	&	0.97	&	0.90	&	0.93	&	0.95	\\
			\bottomrule[1.2pt]
		\end{tabular}
	}
\end{table*}

\begin{table}[htb]
	\fontsize{8pt}{10pt}\selectfont
	\centering
	\captionsetup{justification=justified, singlelinecheck=false}
	\caption{Summary of the XCOR and PCC results for treadmill walking in the K2MUSE dataset and overground walking in the comparison dataset.}
	
	\label{tab:Comparison_LG}
	\renewcommand{\arraystretch}{1.2}
	\begin{tabular}{c l c c c}
		\toprule[1.2pt]
		\multirow{2}{*}{\textbf{Dataset}} & \multirow{2}{*}{\textbf{Metric}} & \multicolumn{3}{c}{\textbf{Level-ground walking}} \\
		\cmidrule(lr){3-5}
		& & \textbf{Ankle} & \textbf{Knee} & \textbf{Hip} \\
		\midrule[1.2pt]
		
		\multirow{2}{*}{\citet{Ramp_Stair}} 
		& XCOR & 0.89 & 0.97 & 0.91 \\
		\cline{2-5}
		& PCC  & 0.80 & 0.91 & 0.98 \\
		\midrule
		
		\multirow{2}{*}{\citet{KK_HDEMG}} 
		& XCOR & 0.89 & 0.99 & 0.97 \\
		\cline{2-5}
		& PCC  & 0.83 & 0.89 & 0.96 \\
		\midrule
		
		\multirow{2}{*}{\citet{KK_EMG2}} 
		& XCOR & 0.88 & 0.99 & 0.96 \\
		\cline{2-5}
		& PCC  & 0.88 & 0.94 & 0.98 \\
		\bottomrule[1.2pt]
	\end{tabular}
\end{table}

\subsection{Motion Intention Recognition}
To assess the performance of decoding lower limb movements using sEMG and AUS signals, we evaluated the recognition performance of different models in both joint angle prediction and gait phase classification tasks.
For data preprocessing, we applied the following steps.
For the sEMG data, a fourth-order Butterworth bandpass filter (20--500 Hz) was initially applied, followed by a 50 Hz notch filter. 
The filtered sEMG data were then segmented into analysis windows of 300 sample points, with a 200 sample point overlap. 
The sEMG features, including the mean absolute value, waveform length, zero crossings, and slope sign changes, were extracted for each window \citep{EMGFeature}.
For the AUS data, the raw signal of each frame was processed sequentially through time gain compensation, bandpass filtering, envelope detection, and log compression \citep{AUSFatigue}. 
To extract AUS data features, the frames were segmented into a series of windows, each containing 20 sample points. The first and last 20 points were discarded prior to segmentation, as they typically do not contain valuable information, resulting in 48 segments per frame \citep{SMG_TMECH}. 
Two types of features were computed from the AUS data, namely, the MSD feature (mean and standard deviation) and the SFO feature (spatial first-order feature) \citep{AUS_SFO}. For each of the four AUS channels, we extracted two statistics per window for each of 48 segments, yielding a 384-dimensional feature vector for each feature set (i.e., 4$\times$2$\times$48=384).

The calculation formulas for the two AUS features are provided below.
Let $x^{(c)}(n)$ denote the envelope-compressed AUS signal of channel $c$ ($c=1,\dots,4$) within one frame, where $n=1,\dots,N$ represents the depth samples.
After discarding the first and last 20 samples, the remaining signal was partitioned into $K=48$ non-overlapping segments of equal length $L=20$.
For segment $k$ ($k=1,\dots,K$), we denote its samples as $x^{(c)}_{k}(\ell)$ with $\ell=1,\dots,L$.

\emph{(i) MSD feature.}
For each segment, the mean and standard deviation were computed as follows:
\begin{equation}
\begin{aligned}
	\mu^{(c)}_{k}&=
	\frac{1}{L}\sum_{\ell=1}^{L}x^{(c)}_{k}(\ell)\\
	\sigma^{(c)}_{k}&=
	\sqrt{\frac{1}{L-1}\sum_{\ell=1}^{L}\left(x^{(c)}_{k}(\ell)-\mu^{(c)}_{k}\right)^2}
\end{aligned}
\end{equation}
yielding the channel-wise MSD feature vector
\begin{equation}
	\mathbf{f}^{(c)}_{\mathrm{MSD}}
	=\big[\mu^{(c)}_{1},\sigma^{(c)}_{1},\mu^{(c)}_{2},\sigma^{(c)}_{2},\dots,\mu^{(c)}_{K},\sigma^{(c)}_{K}\big]^{\mathsf T}
\end{equation}

\emph{(ii) SFO feature.}
For each segment, a first-order (linear) model was fitted along the depth as follows:
\begin{equation}
	x^{(c)}_{k}(\ell)\approx a^{(c)}_{k}\,\ell+b^{(c)}_{k},\qquad \ell=1,\dots,L
\end{equation}
and the two fitting coefficients $\left(a^{(c)}_{k},\,b^{(c)}_{k}\right)$ were used as the SFO features, obtained by least squares as follows: 
\begin{equation}
	\left(a^{(c)}_{k},b^{(c)}_{k}\right)
	=\arg\min_{a,b}\sum_{\ell=1}^{L}\left(x^{(c)}_{k}(\ell)-(a\ell+b)\right)^2
\end{equation}
This yielded the channel-wise SFO feature vector:
\begin{equation}
	\mathbf{f}^{(c)}_{\mathrm{SFO}}
	=\big[a^{(c)}_{1},b^{(c)}_{1},a^{(c)}_{2},b^{(c)}_{2},\dots,a^{(c)}_{K},b^{(c)}_{K}\big]^{\mathsf T}
\end{equation}

Finally, the four-channel features were concatenated as follows:
\begin{equation}
\begin{aligned}
	\mathbf{f}_{\mathrm{MSD}}
	&=\big[\mathbf{f}^{(1)}_{\mathrm{MSD}};\mathbf{f}^{(2)}_{\mathrm{MSD}};\mathbf{f}^{(3)}_{\mathrm{MSD}};\mathbf{f}^{(4)}_{\mathrm{MSD}}\big]\\
	\mathbf{f}_{\mathrm{SFO}}&
	=\big[\mathbf{f}^{(1)}_{\mathrm{SFO}};\mathbf{f}^{(2)}_{\mathrm{SFO}};\mathbf{f}^{(3)}_{\mathrm{SFO}};\mathbf{f}^{(4)}_{\mathrm{SFO}}\big]
\end{aligned}
\end{equation}

All the sEMG and AUS features were normalized using min--max normalization. To reduce redundancy and increase computational efficiency, we applied principal component analysis (PCA) to each AUS feature set (MSD and SFO). As a result, the dimensionality was reduced from 384 to 200 with PCA (n\_components=200), which is consistent with \citet{SMG_TMECH}.

To evaluate the performance of the sEMG, AUS, and multimodal fusion data, the single-modal features included sEMG, MSD, or SFO features. The multimodal features included different feature fusion combinations: sEMG\&MSD, sEMG\&SFO, and MSD\&SFO features. 
In addition, the resulting sequence of preprocessed sEMG analysis windows and AUS frames can be directly used as inputs to deep learning (DL) models.
For model training, we employed 5-fold cross-validation, where each trial was held out in turn as the test fold.
For each participant, the root mean square error (RMSE) and classification accuracy were computed to evaluate the regression and classification performance.

For the machine learning (ML) models, we considered support vector machines, Gaussian processes, and multilayer perceptrons.
For joint angle prediction, we utilized support vector regression (SVR), Gaussian process regression, and multilayer perceptron (MLP) regression models to predict the joint angles of the left leg during level-ground walking at a speed of 1.0 m/s.
For gait phase classification, we followed the protocol of \citet{LiGuangLin} to partition each full gait cycle into five phases, and implemented three classifiers: a support vector classifier (SVC), a Gaussian process classifier (GPC), and an MLP classifier. Given that the focus was not on algorithmic novelty, all the ML models were implemented via \href{https://scikit-learn.org/stable/index.html}{\textcolor{blue}{scikit-learn library}}.

For the DL model, we followed the standard transformer encoder architecture \citep{vaswani2017attention} and adopted a joint-token multimodal fusion strategy, similar to VisualBERT and UNITER \citep{li2019visualbert, chen2020uniter}. The resulting single-stream encoder serves as a unified baseline for both joint angle regression and gait phase classification.
At each time step $t$, the input includes an sEMG window $\mathbf{x}_t^{e}\in\mathbb{R}^{C\times L_e}$ and an AUS frame $\mathbf{x}_t^{u}\in\mathbb{R}^{C\times L_u}$, where $C=4$, $L_e=300$, and $L_u=1000$. 
Each modality is converted to an $d$-dimensional token through a lightweight one-dimensional convolutional tokenizer, followed by global average pooling and a linear projection, producing $\mathbf{e}_t=f_e(\mathbf{x}_t^{e})\in\mathbb{R}^{d}$ and $\mathbf{u}_t=f_u(\mathbf{x}_t^{u})\in\mathbb{R}^{d}.$
The EMG tokenizer uses three Conv1D blocks with channel sizes of 32, 64, and 128 and kernel sizes of 7, 5, and 3 (padding 3, 2, and 1), each followed by BatchNorm and ReLU layers. 
The AUS tokenizer uses channel sizes of 32, 64, and 128 with kernel sizes of 11, 7, and 5 (padding 5, 3, and 2), followed by BatchNorm and ReLU layers. 
The final 128-channel features are globally pooled and linearly projected to the transformer embedding dimension $d$.
For early fusion, we concatenate the EMG and AUS token streams along the sequence dimension to form joint tokens:
\begin{equation}
	\mathbf{X}=\big[\mathbf{e}_1,\ldots,\mathbf{e}_T,\mathbf{u}_1,\ldots,\mathbf{u}_T\big]\in\mathbb{R}^{2T\times d}.
\end{equation}

We then add learned temporal positional embeddings shared across modalities and learned modality embeddings indicating EMG versus AUS, followed by LayerNorm and dropout layers, which was formulated as follows:

\begin{equation}
	\tilde{\mathbf{X}}=\mathrm{LN}(\mathbf{X}+\mathbf{P}+\mathbf{M}).
\end{equation}

The resulting sequence is processed by a transformer encoder with $L=8$ layers, $H=8$ attention heads, a feedforward width of $4d$, and a dropout rate of 0.1 (with $d=256$ in our experiments). 
For variable-length segments, a key-padding mask is applied by replicating the length--$T$ mask to length $2T$.
Instead of pooling across time, we preserve the per-time-step structure by pairing the encoded EMG and AUS outputs at the same index and concatenating them as follows:
\begin{equation}
	\mathbf{z}_t=\big[\mathbf{h}_t^{e};\mathbf{h}_t^{u}\big]\in\mathbb{R}^{2d},\qquad
	\hat{\mathbf{y}}_t=g(\mathbf{z}_t)
\end{equation}
where $\mathbf{h}_t^{e}$ and $\mathbf{h}_t^{u}$ denote the encoder outputs corresponding to the EMG and AUS token positions, respectively. 
A dropout layer is applied before the task head $g(\cdot)$. 
We train the model by minimizing the masked per-time-step loss over valid indices. 
For evaluation, we reconstruct trial-level predictions via sliding-window inference with overlap averaging to form a continuous output sequence before computing the metrics.
For angle estimation, $g(\cdot)$ is a linear regressor $\mathbb{R}^{2d}\!\to\!\mathbb{R}^{3}$ that produces the hip, knee, and ankle angles at each time step, is trained with the masked mean squared error and is reported with the RMSE. 
For phase classification, $g(\cdot)$ is replaced by a linear classifier $\mathbb{R}^{2d}\!\to\!\mathbb{R}^{K}$ with $K=5$ gait phases, which is trained via the masked cross-entropy and evaluated using the sequence-level accuracy.

\subsubsection{Joint angle prediction.}

\begin{figure*}
	\centering
	{\includegraphics[width=1\linewidth]{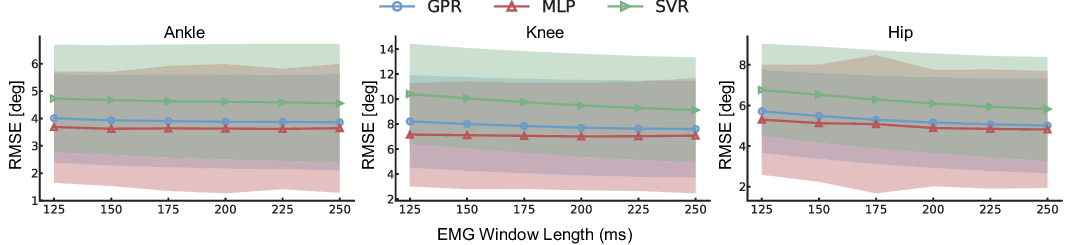}}
	\captionsetup{justification=justified, singlelinecheck=false}
	\caption{Comparison of lower-limb joint-angle estimation errors (RMSE) using sEMG inputs with different window lengths. The curve shows the RMSE trend across window settings, and the shaded region indicates inter-subject variability.}
	\label{fig:DiffWindowRMSEemgFeature}
\end{figure*}

Following common practice in real-time myoelectric decoding, we previously selected a 300-sample analysis window (150 ms at 2000 Hz) to balance estimation stability and latency. Prior studies suggest that very short windows can yield noisier feature estimates, whereas longer windows increase delay and may reduce responsiveness in closed-loop control \citep{godoy2022electromyography, englehart2003robust}. To support this choice for K2MUSE, we additionally performed a window-length sensitivity analysis for sEMG-based joint-angle prediction (Figure~\ref{fig:DiffWindowRMSEemgFeature}). The RMSE decreased as the window length increased, but the improvement largely plateaued at approximately 150 ms, indicating that longer windows provide only marginal gains while incurring additional latency. Therefore, the 300-sample window adopted for sEMG preprocessing and feature extraction is appropriate.

\begin{figure*}
	\centering
	{\includegraphics[width=1\linewidth]{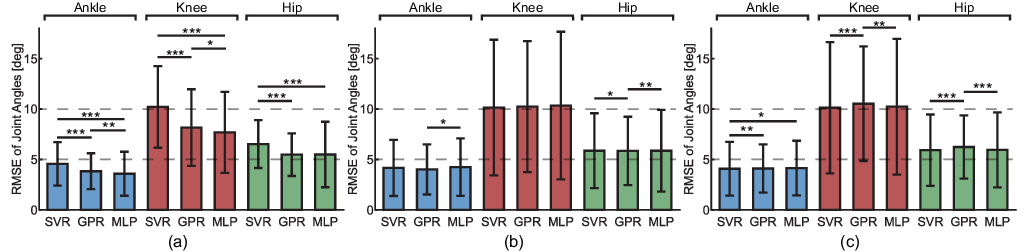}}
	\captionsetup{justification=justified, singlelinecheck=false}
	\caption{Angle estimation regression results with single-modality inputs for different ML models. (a), (b), and (c) show the $RMSE$ obtained with sEMG, MSD, and SFO features as inputs, respectively. The height of the bars represents the mean value across all participants, whereas the error bars indicate the standard deviation.}
	\label{fig:RegressionMetricsSinglemodal}
\end{figure*}

Figure~\ref{fig:RegressionMetricsSinglemodal}~(a)--(c) shows the RMSE results for predicting the ankle, knee, and hip joint angles on the basis of different single features.
For predictions based on the sEMG features, the MLP achieved the best performance, with average RMSE values of 3.58$^\circ$, 7.69$^\circ$, and 5.49$^\circ$ for the ankle, knee, and hip joints, respectively. 
There was no significant difference in the hip-joint RMSE between the MLP and GPR models. However, the MLP showed a significant advantage in all the other comparisons ($p \textless 0.05$).
With respect to the predictions based on the MSD features, all three models exhibited similar performance, with average RMSE values of approximately 4.1$^\circ$, 10.2$^\circ$, and 5.8$^\circ$ for the ankle, knee, and hip joints, respectively. 
Statistical analysis revealed no significant differences among the three models for the knee joint. For the hip joint, significant differences were observed between the SVR and GPR models and between the GPR and MLP models ($p \textless 0.05$). With respect to the ankle joint, the GPR results differed significantly from the MLP results ($p \textless 0.05$).
In the prediction based on SFO features, the SVR model showed a slight advantage, yielding average RMSE values of 4.08$^\circ$, 10.12$^\circ$, and 5.91$^\circ$ for the ankle, knee, and hip joints, respectively.
Statistical tests revealed significant differences between the SVR and GPR models and between the SVR and MLP models for the ankle joint. For the knee joint, significant differences were observed between the SVR and GPR models and between the GPR and MLP models. With respect to the hip joint, the SVR results differed significantly from the GPR results, and the GPR results differed significantly from the MLP results (all $p \textless 0.05$).

\begin{figure*}
	\centering
	{\includegraphics[width=1\linewidth]{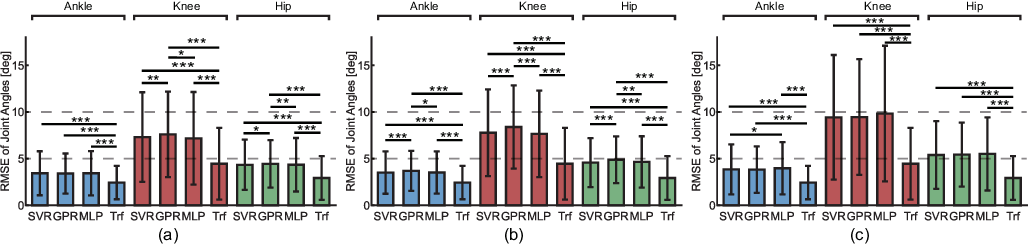}}
	\captionsetup{justification=justified, singlelinecheck=false}
	\caption{Angle estimation regression results with multimodal inputs for different models. (a), (b), and (c) show the RMSE results for the ML models when sEMG\&MSD, sEMG\&SFO, and MSD\&SFO feature combinations are used as inputs, respectively, together with the RMSE of the transformer model under the corresponding multimodal input settings. The height of the bars represents the mean value across all participants, whereas the error bars indicate the standard deviation. `Trf' is an abbreviation for the transformer.}
	\label{fig:RegressionMetricsMultiModal}
\end{figure*}

Figure~\ref{fig:RegressionMetricsMultiModal}~(a)--(c) shows the RMSE results of different models under various multimodal input configurations.
Compared with the use of single-modality features, the performance of the models with the fusion of the sEMG\&MSD and sEMG\&SFO features improved, resulting in smaller RMSE values.
The performance enhancement for the sEMG\&MSD and sEMG\&SFO fusion features was particularly pronounced for the SVR and GPR models, with the average RMSE values for the ankle, knee, and hip joint angles reduced to approximately 3.4$^\circ$, 7.3$^\circ$, and 4.3$^\circ$, respectively.
For predictions based on the sEMG\&MSD features, no significant differences were observed among the SVR, GPR, and MLP models for the ankle joint. For the knee and hip joints, significant differences were found between the SVR and GPR models and between the GPR and MLP models ($p\textless0.05$). 
For predictions based on the sEMG\&SFO features, significant differences were consistently observed between the SVR and GPR models and between the GPR and MLP models across all three joints ($p\textless0.05$). 
For predictions based on the MSD\&SFO features, a significant difference was observed between the SVR and MLP models for the ankle joint ($p\textless0.05$), whereas no significant differences were found among the SVR, GPR, and MLP models for the knee and hip joints ($p\textless0.05$).
The combination of the MSD\&SFO features did not improve model performance, yielding RMSE values similar to those obtained from single MSD or SFO features.
Therefore, combining different features of the same modality did not yield significant performance improvements. 
However, the fusion of the sEMG and AUS signals, which incorporates both action potential information and thickness variation data from muscle contraction, demonstrates the potential of multimodal fusion for enhanced intent recognition.
These results demonstrate the applicability of our dataset (K2MUSE) for lower limb joint angle estimation.
For ease of comparison, we added the results of the transformer model using sEMG and AUS data as multimodal inputs to each subplot in Figure~\ref{fig:RegressionMetricsMultiModal}.
Overall, the transformer achieved low regression errors across all three joints, with RMSEs of 2.43$\pm$1.78$^\circ$, 4.45$\pm$3.84$^\circ$, and 2.93$\pm$2.34$^\circ$ for the ankle, knee, and hip joints, respectively.
Statistical analysis revealed that the transformer significantly outperformed all the conventional ML models when different fused feature sets ($p\textless0.05$) were used for all three joints.
These results suggest that end-to-end sequence modeling with multimodal fusion can better exploit cross-modal temporal information, leading to increased accuracy and robustness.

\begin{figure*}
	\centering
	{\includegraphics[width=1\linewidth]{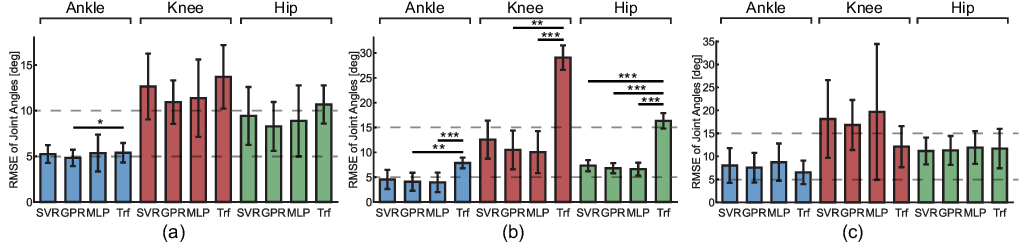}}
	\captionsetup{justification=justified, singlelinecheck=false}
	\caption{Angle estimation regression results of different models under non-ideal conditions. (a), (b), and (c) show the $RMSE$ results for the electrode shift, muscle fatigue, and interday difference experiments, respectively. The height of the bars represents the mean value across all participants, whereas the error bars indicate the standard deviation. `Trf' is an abbreviation for the transformer.}
	\label{fig:RegressionMetricsNonideal}
\end{figure*}

We evaluated recognition performance under three non-ideal conditions using ML models with sEMG features as inputs and the same transformer model described above. 
To effectively demonstrate the impact of these non-ideal conditions, we implemented testing protocols that deviated from conventional cross-validation approaches.
For the electrode shifts, data from the initial position were used as the training dataset, while data from the other four shift positions served as the testing dataset. 
In the case of muscle fatigue, data from the first trial were designated as the training set, with data from the remaining nine trials used as the testing dataset. 
For inter-day differences, data from the ideal condition experiments served as the training dataset, while data collected on a separate day were used as the testing dataset.
The obtained angle prediction results are illustrated in Figure~\ref{fig:RegressionMetricsNonideal}~(a), (b), and (c).
Taking the MLP as an example, the RMSEs for the ankle, knee, and hip joints under electrode shift, muscle fatigue, and inter-day difference conditions are 5.36$^\circ$, 11.36$^\circ$, and 8.88$^\circ$; 3.96$^\circ$, 10.05$^\circ$, and 6.61$^\circ$; and 8.73$^\circ$, 19.67$^\circ$, and 11.92$^\circ$, respectively.
Compared with the data in Figure~\ref{fig:RegressionMetricsSinglemodal}~(a), the RMSEs under the three non-ideal conditions increase to varying extents, demonstrating the negative impact of non-ideal conditions on recognition performance. This observation is consistent with the analysis of the MAV features presented in Section~\ref{NonidealAnalysis}.
In addition, Figure~\ref{fig:RegressionMetricsNonideal} indicates that under the three non-ideal conditions, the transformer does not consistently outperform the ML models in terms of the regression error and even shows greater degradation for certain joints and scenarios.
This behavior is closely tied to our evaluation protocol, where the model is trained on a single source domain and tested on target domains with substantial distribution shifts.
Under such strong domain shifts, transformers may be more susceptible to relying on source domain-specific signal features, which do not transfer well across nonideal conditions. Consequently, cross-domain generalization becomes the dominant bottleneck, potentially outweighing the representational capacity of the transformer. Overall, these results suggest that incorporating transfer learning techniques may further increase robustness and better utilize the transformer's potential in nonideal scenarios \citep{prahm2019counteracting}.

\subsubsection{Gait Phase Classification.}

\begin{figure*}
	\centering
	{\includegraphics[width=1\linewidth]{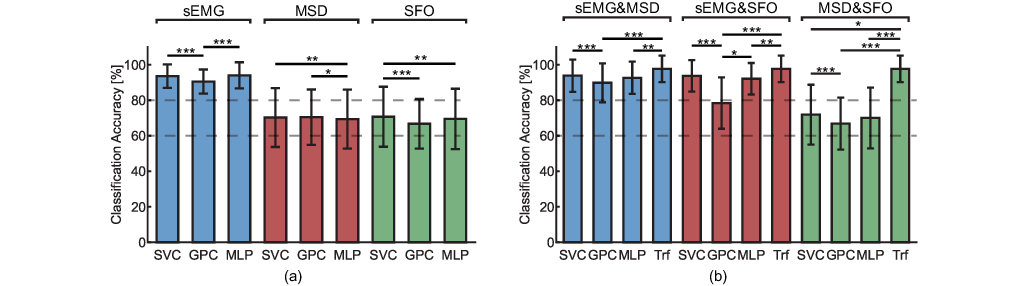}}
	\captionsetup{justification=justified, singlelinecheck=false}
	\caption{Gait phase classification results of different models under different input settings. (a) Classification accuracy of the ML models using different single-modal feature inputs. (b) Classification accuracy of the ML models using fused feature inputs, together with the accuracy of the transformer under the corresponding multimodal inputs. The height of the bars represents the mean value across all participants, whereas the error bars indicate the standard deviation. `Trf' is an abbreviation for the transformer.}
	\label{fig:ClassificationMetrics}
\end{figure*}

Figure~\ref{fig:ClassificationMetrics} shows the gait phase classification accuracies under different input settings. As shown in Figure~\ref{fig:ClassificationMetrics}(a), under single-modal inputs, the ML models based on the sEMG features achieved high performance, reaching accuracies of 94.05$\pm$7.39\% (MLP), 93.60$\pm$6.62\% (SVC), and 90.55$\pm$6.78\% (GPC). 
In contrast, the accuracy decreased markedly when AUS-derived features were used. 
Specifically, the MSD yielded accuracies of approximately 69--71\% (69.39$\pm$16.60\%--70.49$\pm$15.60\%), and the SFO yielded accuracies of approximately 67--71\% (66.76$\pm$13.95\%--70.73$\pm$16.90\%), with substantially greater dispersion, 
suggesting more pronounced inter-subject differences.
Statistical analysis indicated significant differences among the models for each input setting ($p\textless0.05$). Specifically, for the sEMG signals, significant differences were detected between the MLP and GPC models and between the SVC and GPC models. With respect to the MSD, the MLP results differed significantly from both the SVC and GPC results. With respect to the SFO, significant differences were observed between the MLP and SVC models and between the SVC and GPC models.
As shown in Figure~\ref{fig:ClassificationMetrics} (b), under feature-level fusion inputs, the performance gains of the ML models were largely attributable to the inclusion of sEMG data. 
With the sEMG\&MSD combination, the MLP and SVC models achieved accuracies of 92.65$\pm$9.09\% to 93.84$\pm$9.10\%, and with the sEMG\&SFO combinations, they reached accuracies of 92.13$\pm$8.88\% to 93.73$\pm$8.87\%. 
In contrast, when only AUS features were fused (MSD\&SFO), the accuracy remained at approximately 70\% (70.04$\pm$17.10\% to 71.91$\pm$16.85\%), indicating the limited discriminative power for gait-phase identification when relying solely on AUS-derived features. 
By comparison, the transformer achieved higher and more stable accuracy across all three multimodal combinations (97.67$\pm$7.47\%), and the overall model differences were significant ($p\textless0.05$). 
These results suggest that gait-phase information is more salient in sEMG data, whereas AUS features alone exhibit greater inter-subject variability. Moreover, end-to-end multimodal temporal modeling can better exploit complementary cross-modal information, yielding more stable and higher classification performance than handcrafted feature fusion.

\subsection{Statistical analysis}
All the statistical analyses were performed with SPSS 26.0 (IBM SPSS Statistics, USA) to assess the significance of the differences across all the metrics. 
Data normality was first evaluated using the Shapiro--Wilk test. 
For normally distributed data, one-way analysis of variance (ANOVA) and paired t tests were used to examine significant differences among conditions.
Otherwise, one-dimensional Friedman's ANOVA and the Wilcoxon signed-rank test were applied. 
For multiple comparisons, $p$ values were adjusted using the Holm--Bonferroni correction. 
In this paper, the $*$, $**$, and $***$ symbols in the figures indicate significant differences at the $p\textless0.05$, $p\textless0.01$, and $p\textless0.001$ levels, respectively.

\subsection{Data limitations}
Given that the dataset involved multiple acquisition systems and extensive data collection sessions, only data from female participants walking at 1.0 m/s on different ramps were included. 
Owing to recruitment limitations, the participant diversity in this dataset is restricted, preventing the data from fully representing the entire population. 
During data validation, certain trials were flagged as invalid because of acquisition anomalies. These primarily included cross-plate strikes, where a foot contacted multiple force plates or the gap between them, which violates the single-plate contact assumption and prevents accurate inverse-dynamics computation. Trials were also rejected when critical marker occlusions prevented reliable trajectory reconstruction in Vicon Nexus. To ensure high data quality, invalid trials were discarded and reacquired immediately using the standard protocol.

Although K2MUSE includes data spanning multiple walking speeds, ramp inclines, and representative non-ideal sensing conditions, most trials were collected under controlled treadmill locomotion settings. This design increases repeatability, enables precise control of the speed and incline, and supports high-fidelity synchronization across kinematics, kinetics, sEMG, and AUS data. These properties are important for benchmarking and for developing control-oriented models.
However, treadmill walking does not fully reflect the ecological complexity of real-world ambulation. Daily locomotion is often self-paced and non-stationary and may include turning, obstacle negotiation, uneven terrain walking, unexpected perturbations, and concurrent tasks \citep{TCN_Nature}. Such factors can introduce distribution shifts in both biomechanics and wearable signals, which can limit the direct transfer of models or controllers trained only on treadmill data to out-of-lab rehabilitation robotics scenarios. 
In addition to the environmental constraints discussed above, K2MUSE currently does not include recordings from rehabilitation patients, such as individuals post-stroke, with neurological disorders, or with limb loss. Although we expanded the cohort to include older adults, clinical populations may exhibit markedly different kinematic synergies, neuromuscular coordination patterns, compensatory strategies, and signal characteristics that are not fully captured by able-bodied participants or aging-related adaptations. Therefore, algorithms and controllers developed using K2MUSE should not be assumed to generalize directly to clinical populations without further validation.

We therefore position K2MUSE as a robust normative baseline and benchmarking resource for method development. For deployment beyond controlled laboratory settings, additional overground data collection, domain adaptation strategies, and, when appropriate, subject- and context-specific calibration may be required. Translation to clinical use will likely require dedicated patient data collection, appropriate safety protocols, and cross-population adaptation studies in future releases.

\section{Potential applications}
\label{PotentialApplications}

Our dataset encompasses a diverse range of variables, comprehensively covering multiple terrains and acquisition conditions. 
In biomechanical analysis, these varied kinematic, dynamic, and physiological data (sEMG, AUS) enable a comprehensive analysis of body movement performance from the perspective of motion mechanisms, particularly through the integration of muscle synergy \citep{MuscleSynergy, SynergyKnee, mehryar2016muscle}. This rich dataset also provides valuable data support for big data-driven methods to uncover powerful solutions for intent decoding \citep{AUS_Torque, EMGEstimation, javed2024multi}.
Additionally, the extensive human data in our dataset enable a more comprehensive evaluation of robotic performance \citep{mahmood2022evaluation}.
Currently, rehabilitation robots, such as exoskeletons, rely primarily on straps and similar mechanisms for physical interaction. However, the underlying interaction mechanism and mechanical coupling often result in an expected assistance effect that is not fully transmitted, leading to an efficiency gap. By leveraging our dataset, this efficiency gap can be better analyzed and quantified, providing a solid foundation for optimizing exoskeleton design and improving overall assistive performance.

\begin{figure}
	\centering
	{\includegraphics[width=0.9\columnwidth]{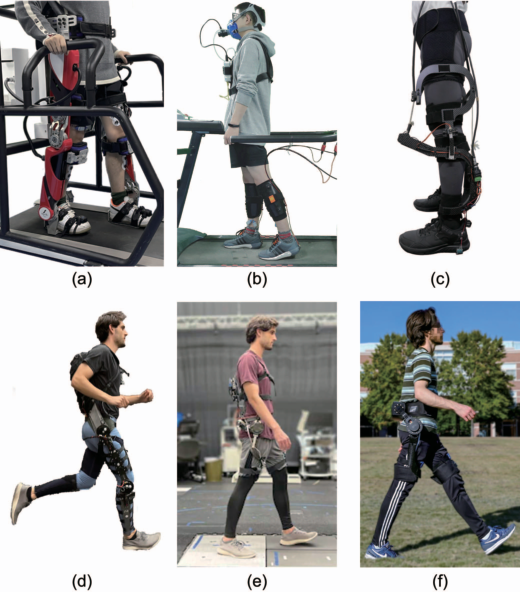}}
	\captionsetup{justification=justified, singlelinecheck=false}
	\caption{Innovative exploration of various exoskeleton designs and control strategies: (a) hip-knee-ankle exoskeleton with series elastic actuators, adapted from \cite{Chen_TRO}; (b) soft exosuit, adapted from \cite{Tan_TRO}; (c) ligament-inspired knee exoskeleton, adapted from \cite{Zhao_RAL}; (d) task-agnostic exoskeleton control, adapted from \cite{TCN_Nature}; (e) unified exoskeleton control, adapted from \cite{TCN_SR}; (f) sim2real framework for the exoskeleton, adapted from \cite{Simulation}.}
	\label{fig:Potential}
\end{figure}

In bionic design for robots, as shown in Figure~\ref{fig:Potential}~(a)--(c), exoskeletons are evolving toward a rigid--flexible coupling approach, progressing from joint assistance with elastic actuators \citep{Chen_TRO} to muscle assistance via tendon-driven mechanisms \citep{Tan_TRO} and further toward bioinspired optimized designs \citep{Zhao_RAL}.
To increase their effectiveness, the system design of rehabilitation robots should align with actual biomechanics, ensuring that actuation effects conform to physiological principles. 
The K2MUSE dataset not only provides comprehensive data on joint and limb movements but also captures physiological changes in muscle contraction, enabling the development of more ergonomic and biomimetic robotic designs.
In terms of controller design for robots, adaptable control strategies are needed to handle variations in terrains and movement patterns. As shown in Figure~\ref{fig:Potential}~(d), an end-to-end controller developed based on our dataset can dynamically adjust to changing motion tasks while ensuring scalability \citep{TCN_Nature}. 
As shown in Figure~\ref{fig:Potential}~(e) and (f), without relying on manual tuning or predefined control laws, these controllers can be generalized to a wide range of tasks as the dataset expands, increasing their adaptability and robustness in real-world applications \citep{TCN_SR, Simulation}.

To substantiate these potential applications, we validate the K2MUSE dataset through several representative wearable-robot tasks and complement the evaluation with quantitative robotics benchmarks. Specifically, we demonstrate how the K2MUSE dataset can be leveraged for mechanical design, quantitative metric-based assessment, and exoskeleton assistance evaluation. Through these task-oriented examples and quantitative evaluations, we further clarify that the K2MUSE dataset not only can be used in biomechanical analysis and intent decoding but also provides direct and verifiable support for rehabilitation robotics at the design, analysis, and control levels.

\subsection{Dataset for mechanical design}

Wearable robots are representative human--robot coupled systems, and the choice of the assistance strategy, mechanical architecture, and joint configuration in wearable robots is often guided by biomimetic insights into human physiology and walking function \citep{yang2026knee}. Consequently, the mechanism design should systematically account for key biomechanical factors, including the allocation of degrees of freedom, structural form, joint range of motion (ROM), joint type, actuation and transmission principles, and the human--robot interface. These considerations are rooted in functional anatomy and biomechanical analysis \citep{firouzi2025biomechanical}. The K2MUSE dataset provides joint-level variables measured using the motion capture system and force plates, such as joint angles, joint moments, and joint power. These variables can be used to determine the required device ROM and to align the kinematic limits of the robot with the human ROM. This alignment helps reduce the risk of overconstraining the model and increases operational safety \citep{boone1979normal}.	 In addition, the kinematic and kinetic measurements in the K2MUSE dataset can serve as reference data for assessing the fidelity of digital twin musculoskeletal models and skeletal dynamics simulations. For example, outputs computed in OpenSim or AnyBody can be compared with the dataset references to evaluate whether the digital model captures human movement with sufficient reliability \citep{troster2026person}. The K2MUSE dataset also includes anthropometric measurements and experimental parameters, such as height, body mass, knee width, ankle width, and leg length. These records can inform adjustable sizing mechanisms and interface layout design, thereby improving joint axis alignment and wearable fit \citep{chen2024systematic}.
Furthermore, the K2MUSE dataset provides synchronized kinematic, kinetic, sEMG, and AUS data across a wide range of conditions. These include changes in speed and incline, as well as data collected under non-ideal conditions such as muscle fatigue, electrode shifts, and inter-day differences. With these features, the dataset can support mechanism-constrained modeling and digital twin studies for rehabilitation robotics by enabling systematic evaluation of model robustness and domain generalization under realistic perturbations. Moreover, subject-scaled musculoskeletal simulations can be used to derive latent indicators, such as ligament or tendon internal forces and cumulative damage measures, which can act as supervisory signals. These signals can be leveraged to train efficient deep surrogate models for near real-time risk inference, offering a pathway toward safety-constrained assistive force regulation and prescription optimization that avoids overload \citep{xu2025data}.

\subsection{Dataset for quantitative metrics}

Quantitative analysis of human gait and lower-limb motion is critical for rehabilitation assessment, human--robot interaction studies, and the evaluation of wearable-robot assistance.
In human--robot coupled systems such as lower-limb exoskeletons, multimodal sensing approaches have been used in prior studies to capture the information needed for such evaluations \citep{chen2016toward,wang2020evaluation,yang2024ultrasound}. The establishment of quantitative metrics to characterize walking ability enables clear tracking of changes in spatiotemporal gait characteristics throughout training.
To demonstrate the utility of the K2MUSE dataset for human motion analysis and quantitative assessment, we adopted the three metrics proposed in \citet{chen2025manifesting} and conducted computational and comparative analyses using the K2MUSE dataset. This evaluation shows that the K2MUSE dataset supports reusable pipelines for quantitative metric computation and enables movement-state analysis across different populations. The metrics are defined as follows.
\begin{enumerate}
	\item Gait Restoration (GR): This metric quantifies differences between the observed joint motion and the template gait pattern.
	\item Phase-Deviation Weighting (PDW): This metric is used to evaluate the temporal coordination and phase synchronicity of joint movements within the gait cycle.
	\item Multijoint Coordination (MJC): This metric is used to assesse synergistic relationships among lower limb joints through configuration space trajectory analysis.
\end{enumerate}

Within the K2MUSE dataset, participants were divided into three subsets: P01--P20 were used as the template set to establish a nominal reference; P21--P30 formed the healthy group; and P32--P39, P41, and P42 constituted the elderly group. Owing to an insufficient number of trials and the absence of retro-reflective markers, data from participants P31 and P40 were excluded from the analysis.
On the basis of this split, we computed the three metrics for each trial across the healthy and elderly groups to compare population-level differences in movement patterns and phase-related characteristics under the same walking task. In all the analyses, the ideal-condition level-ground walking data at 1.0 m/s were used.

\begin{figure}
	\centering
	{\includegraphics[width=1\columnwidth]{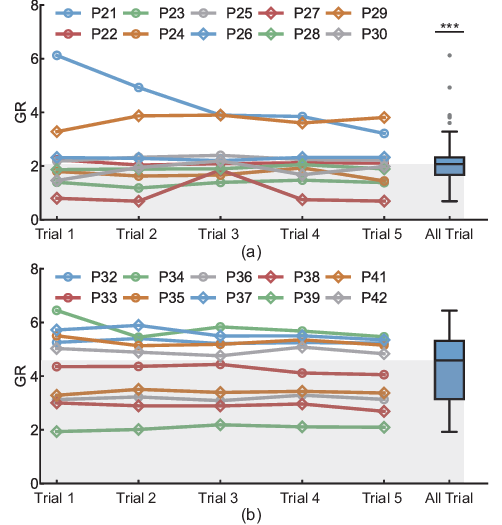}}
	\captionsetup{justification=justified, singlelinecheck=false}
	\caption{Gait restoration results for the 10 participants in (a) the healthy group and (b) the elderly group. The gray shading indicates the region at or below the boxplot median. $***$ indicates a significant difference between the healthy and elderly groups using summarized data across all trials.}
	\label{fig:MetricGR}
\end{figure}

The GR metric, defined as a PCA-based comprehensive distance computed from standardized and decorrelated gait variables, provides an interpretable measure of how closely an individual's gait pattern matches the template distribution, with a larger GR value indicating greater deviation from the nominal reference. Figure~\ref{fig:MetricGR} presents the GR values for each participant and trial and provides the summarized data for all trials. Unlike hemiplegic rehabilitation scenarios in which the GR may progressively approach the nominal range over within-session stages, both cohorts in the K2MUSE dataset exhibited strong inter-trial consistency for the GR value. Within each cohort, comparisons across the five trials revealed no significant trial effect (both $p\textgreater0.05$), indicating that the GR remained stable across repeated trials in both age groups. Despite these stable within-cohort trends, the two groups were clearly separated. When the GR values were summarized across all the trials, compared with the healthy group, the elderly group exhibited significantly higher GR values ($p\textless0.001$). Specifically, the aggregated GR values were 2.26$\pm$1.07 for the healthy group and 4.26$\pm$1.25 for the elderly group. Overall, these results show that age-related differences in gait patterns can be quantitatively captured by the GR metric using a shared template reference, while the absence of within-group trial effects supports the repeatability and benchmarking value of the K2MUSE dataset for quantitative metric validation.

\begin{figure}
	\centering
	{\includegraphics[width=1\columnwidth]{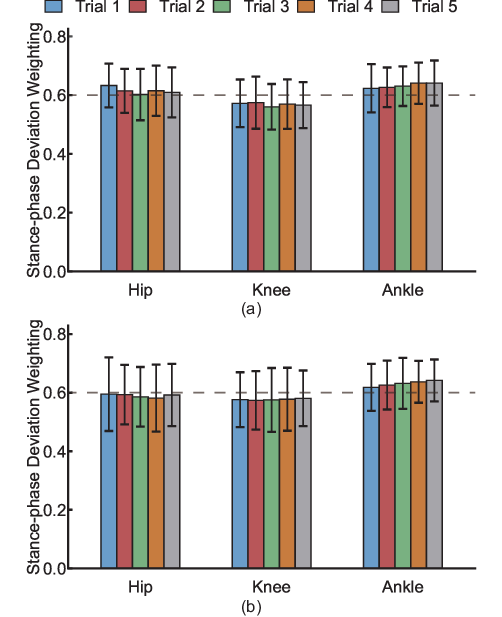}}
	\captionsetup{justification=justified, singlelinecheck=false}
	\caption{Phase-deviation weighting results during the stance phase for the (a) healthy and (b) elderly groups, labeled by trial order.}
	\label{fig:MetricPDW}
\end{figure}

Furthermore, we quantified the phase deviation weighting for the three joints during the stance phase, as shown in Figure~\ref{fig:MetricPDW}. 
Overall, the PDW value exhibited high trial-to-trial consistency in both groups, whereas the degree of stability and dispersion varied across the joints. 
For the hip joint, the mean PDW values across trials 1--5 were approximately 0.60--0.63 in the healthy group and 0.58--0.59 in the elderly group. Within-group comparisons revealed no significant trial effect in either cohort (both $p\textgreater$0.05), indicating that the stance-phase hip PDW value remained stable across repeated trials under the selected walking condition. 
For the knee joint, the mean PDW values were slightly lower than those for the hip and ankle joints, ranging from approximately 0.56--0.57 in the healthy group and 0.57--0.58 in the elderly group. No significant differences across trials were observed within either cohort (both $p\textgreater$0.05), suggesting stable knee PDW values across repeated trials. 
With respect to the ankle joint, the PDW value remained relatively high in both groups and varied minimally across trials, with mean values of approximately 0.62--0.64 for both cohorts. Similarly, within-group analyses revealed no significant trial effect (both $p\textgreater$0.05), supporting the good repeatability of the stance-phase ankle PDW values. 
When the data across all five trials were compared between groups, the PDW value did not significantly differ between the healthy and elderly groups for any of the three joints (all $p\textgreater$0.05). This finding is consistent with the substantial overlap of the bar means and error bars in Figure~\ref{fig:MetricPDW}, indicating that, under the selected conditions in the K2MUSE dataset, the stance-phase PDW value is not sensitive to age-group differences and exhibits a stable distribution across trials.

\begin{figure}
	\centering
	{\includegraphics[width=1\columnwidth]{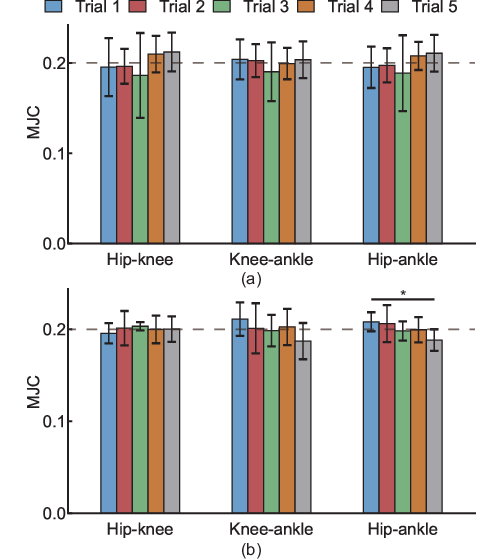}}
	\captionsetup{justification=justified, singlelinecheck=false}
	\caption{Comparison of MJC metrics between (a) the healthy and (b) elderly groups for the hip--knee, knee--ankle, and hip--ankle configurations.}
	\label{fig:MetricMJC}
\end{figure}

To quantify inter-joint interactions from the perspective of spatial coordination, we computed the multijoint coordination metric and analyzed three joint configurations: adjacent joint pairs (hip and knee, knee and ankle) and a distal combination (hip and ankle). Consistent with the reference study by \citet{chen2025manifesting}, in which percentage normalization was applied to the accumulated values, we accumulated the MJC value over multiple gait cycles within each trial and then normalized the accumulated values across Trials 1 to 5 to obtain $\bar{e}_c^s$, as shown in Figure~\ref{fig:MetricMJC}. 
Overall, the $\bar{e}_c^s$ values for both groups clustered at approximately 0.2 across all three configurations, indicating the strong trial-to-trial consistency of the MJC metric. 
In the healthy group, no significant differences were observed across Trials 1 to 5 for any configuration (all $p\textgreater0.01$), suggesting that the MJC distribution remained stable across repeated trials. 
In the elderly group, no significant trial effects were found for the hip and knee or for the knee and ankle configurations (both $p\textgreater0.01$). 
Notably, the elderly group showed an overall significant effect across trials for the hip and ankle configuration ($p\textless0.05$), and post hoc comparisons indicated that Trial 1 differed from Trial 5 after Bonferroni correction ($p_{adj}=0.0236$). However, the mean values in Figure~\ref{fig:MetricMJC} remained close to 0.2, with only modest fluctuations, suggesting that this significance likely reflects a small trial order-related distribution shift rather than a large structural change in coordination. 
When the data across the five trials and cohorts were compared, no significant between-group differences were observed for any configuration. These results suggest that under the selected walking condition in the K2MUSE dataset, the MJC metric primarily serves as a stable baseline across trials, with limited discriminative power between healthy and elderly groups, while still providing a useful measure to validate the repeatability of multi-joint coordination quantification metrics in repeated experiments.

\subsection{Dataset for exoskeleton assistance}

To validate the usability of the proposed K2MUSE dataset for robotic control, we conducted assistive experiments using a soft exoskeleton previously developed by our team \citep{zhang2025modular}. 
In all the experiments, hip-joint assistance was implemented, which served as a control example to demonstrate the effectiveness and practical applicability of the K2MUSE dataset in control-oriented scenarios.
In addition, the exoskeleton platform was equipped with extended Bowden cables and self-locking casters, enabling convenient mobility while maintaining stability when needed, thereby supporting assistive experiments in outdoor environments.

\begin{figure}
	\centering
	{\includegraphics[width=1\columnwidth]{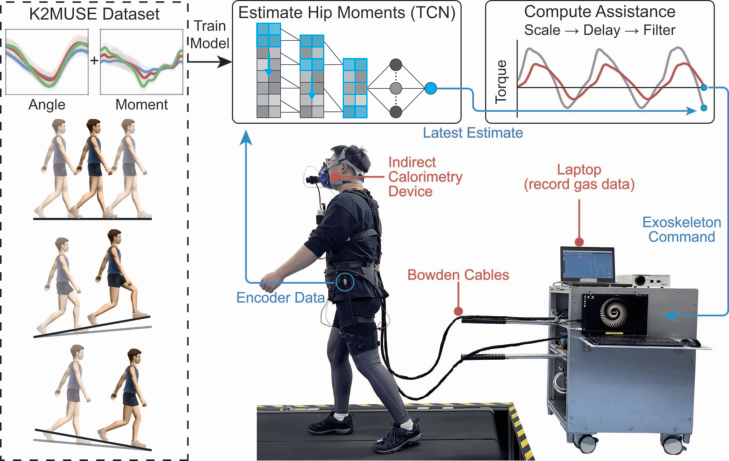}}
	\captionsetup{justification=justified, singlelinecheck=false}
	\caption{Framework for joint moment control. A TCN trained on the K2MUSE dataset estimates the user's hip joint moment from encoder measurements collected at the hip joint. The mid-level control layer processes the instantaneous moment estimates produced by the TCN following the procedure described by \citet{TCN_SR}.}
	\label{fig:ControlFramework}
\end{figure}

Following \citet{TCN_SR}, we adopt an end-to-end control framework in which a DL model is used to estimate the wearer's hip joint moment online solely from the real-time hip joint angle. 
The estimated moment is directly used as the control command to provide natural assistance during walking. The overall pipeline is illustrated in Figure~\ref{fig:ControlFramework}.
Specifically, a temporal convolutional network (TCN) is first trained using the kinematic and kinetic measurements collected during multiple walking tasks in the constructed K2MUSE dataset. 
The trained model is then deployed on an embedded processor (Jetson Orin NX, NVIDIA) for online inference, where the real-time hip joint angle is acquired via an encoder and fed into the network. 
To reduce inference latency, the pretrained model is converted into a TensorRT engine during deployment. 
The estimated joint moment is further processed through scaling, delay and filtering and then mapped to the assistance input force profile of the soft exoskeleton.
At the low-level control layer, an admittance controller and a PD controller generate the motor velocity commands to execute assistance functions.

\begin{figure}
	\centering
	{\includegraphics[width=0.9\columnwidth]{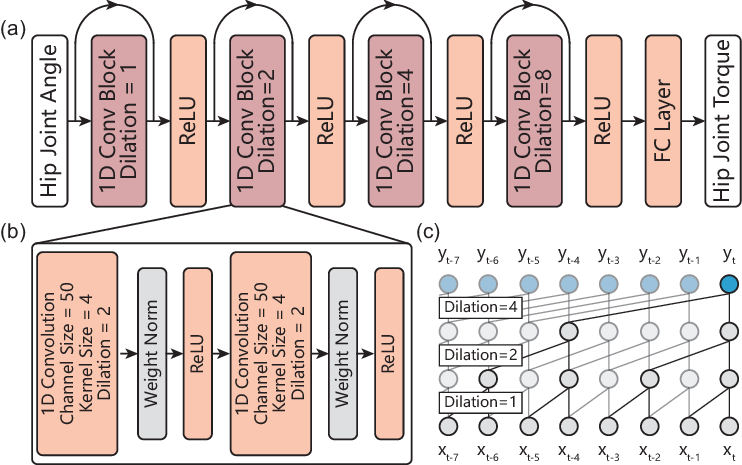}}
	\captionsetup{justification=justified, singlelinecheck=false}
	\caption{TCN structure for hip moment estimation.}
	\label{fig:TCN_structure}
\end{figure}

The adopted TCN architecture follows \citet{bai2018empirical}, as illustrated in Figure~\ref{fig:TCN_structure}. The network is composed of four residual blocks stacked sequentially. 
Each residual block contains two layers of one-dimensional causal convolutions.
The output of each convolutional layer is processed by weight normalization and an ReLU activation function, and a residual connection is applied at the end of the block to facilitate stable training and feature propagation. 
The numbers of channels in the four blocks are 16, 32, 64, and 128. The kernel size is 4, and the dropout rate is 0.3.
Given that the modular exoskeleton provides unilateral hip assistance, the TCN input sequence is derived from the left hip joint angle, and the output labels correspond to the left hip flexion/extension moment. 
Accordingly, the TCN was trained using the PyTorch framework on a workstation equipped with an NVIDIA GeForce RTX 4070 GPU. We used the Adam optimizer with mean squared error as the loss function and set the learning rate to 0.001.

\begin{figure}
	\centering
	{\includegraphics[width=0.9\columnwidth]{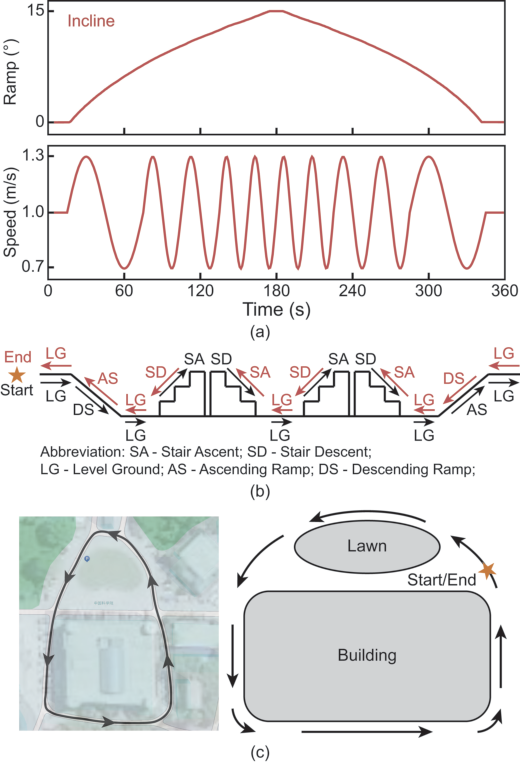}}
	\captionsetup{justification=justified, singlelinecheck=false}
	\caption{Experimental setup for three assisted walking conditions. (a) Treadmill assistance with dynamically varying speeds and inclines. (b) Complex-terrain assistance across level ground, ramps, and stairs. (c) Outdoor overground assistance around a campus building and lawn.}
	\label{fig:AssistScenario}
\end{figure}

The assistive evaluation comprised three walking conditions, as shown in Figure~\ref{fig:AssistScenario}.
Videos of the experiments are available at \href{https://k2muse.github.io/datasets/index#videos-of-experiments}{\textcolor{blue}{K2MUSE/EXOSKELETON ASSISTANCE EXPERIMENT}}.
For each condition, participants wore the exoskeleton and performed two trials: assistance-on and assistance-off. 
A total of nine participants who also participated in the K2MUSE data collection protocol were recruited. 
Their heights and body masses ranged from 173.00--186.70 cm and 64.30--84.75 kg, respectively, covering a broad spectrum of anthropometrics.

\begin{enumerate}
	\item \textbf{The treadmill assistance component} followed the protocol in \citet{lee2024ai}. Using the predefined profiles in Figure~\ref{fig:AssistScenario}(a), participants walked with varying speeds and inclines while wearing the exoskeleton. During the 6-min trial, the incline was held at 0$^\circ$ for the first 25 s, after which the incline command was switched to 15$^\circ$. Once the maximum incline was reached, it was maintained for 20 s; then, the incline returned to 0$^\circ$ and was again held constant for 25 s until the trial ended. The speed changed according to a piecewise sinusoidal profile: in the first half of the experiment, the participant walked for 15 s at a constant speed of 1 m/s; then, the speed varied in a sinusoidal manner with periods of 60 s, 30 s, and 22.5 s for 60 s, 60 s, and 45 s, respectively. In the second half, the changes mirrored those in the first half in reverse order. This condition aimed to examine assistance performance under dynamically changing speeds and inclines.
	\item \textbf{The complex-terrain assistance condition} included level-ground walking, incline walking, decline walking, and stair ascent/descent tasks.
	As illustrated in Figure~\ref{fig:AssistScenario}(b), participants started from a designated point and subsequently completed level-ground walking, descending ramp walking, two rounds of stair ascent/descent walking, and ascending ramp walking before returning to the start along the same route. The route was repeated three times per trial, with an overall duration of approximately 5 min (which varied slightly across participants). This condition was designed to evaluate the assistance performance of the model trained on the K2MUSE dataset under complex and unstructured terrain transitions.
	\item \textbf{The outdoor overground assistance experiments} were conducted on the campus of the Shenyang Institute of Automation, Chinese Academy of Sciences, to assess performance in a real-world walking context. As shown in Figure~\ref{fig:AssistScenario}(c), participants walked a loop around an office building and a lawn area and returned to the starting point. Each trial lasted approximately 6 min (with slight participant-dependent variations).
\end{enumerate}

Metabolic cost was used to quantify the assistance effect. It was estimated from the measured oxygen consumption and carbon dioxide production using the modified Brockway equation \citep{brockway1987derivation}.
Gas exchange data were collected using a portable indirect calorimetry system (Oxycon Mobile, Vyaire Medical, Germany) \citep{Tan_TRO}.
For each walking condition, participants performed 3 min of quiet standing while wearing the exoskeleton both before and after the trial to measure the resting metabolic cost. The metabolic cost for each condition was then obtained by subtracting the resting metabolic cost from the metabolic cost measured during walking.

\begin{figure*}
	\centering
	{\includegraphics[width=1\linewidth]{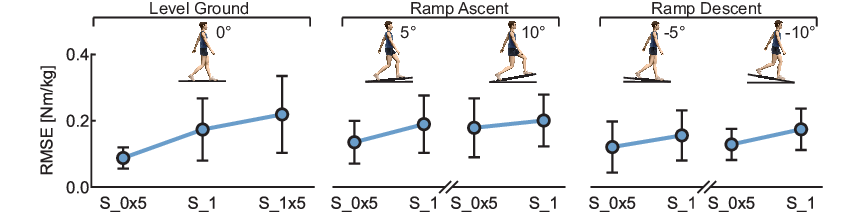}}
	\captionsetup{justification=justified, singlelinecheck=false}
	\caption{Comparison of mean TCN RMSEs across ambulation tasks at different walking speeds.}
	\label{fig:TCN_RMSE}
\end{figure*}

Figure~\ref{fig:TCN_RMSE} shows the offline joint moment regression performance of the trained TCN on the K2MUSE test set across different slopes and walking speeds. Performance is quantified using the root mean square error (RMSE, mean $\pm$ standard deviation).
The tasks include level-ground walking (0$^\circ$), uphill walking (5$^\circ$ and 10$^\circ$), and downhill walking (-5$^\circ$ and -10$^\circ$), with each task performed at multiple speeds. In each panel, the dots denote the mean RMSE, the error bars indicate the inter-subject standard deviation, and the lines connecting results at different speeds illustrate speed-dependent trends. Across all the tasks, the mean RMSE remains below 0.22 Nm/kg, indicating the stable regression performance of the trained TCN as a moment estimator under varying task conditions. The lowest and most consistent errors were obtained under the level-ground walking conditions. In contrast, greater errors were obtained under the incline walking conditions, with increases in the slope magnitude and walking speed.

Table~\ref{tab:Metabolic_Summary} summarizes the metabolic cost across the treadmill, complex-terrain, and outdoor overground walking experiments. Overall, compared with the no-assistance condition, assistance reduced the metabolic cost in all three tasks, with mean values of 6.9868$\pm$0.8350, 3.3893$\pm$0.2534, and 2.9505$\pm$0.6215, respectively. The subject-level results show consistent reductions for all participants in both treadmill and complex-terrain walking, indicating that the assistance yielded stable metabolic benefits under controlled conditions and during more perturbed terrain-transition tasks. 
In the outdoor overground walking experiment, the metabolic cost decreased for all participants when assistance was provided. However, the magnitude of the reduction varied across individuals, suggesting heterogeneous responses in real-world contexts.

\begin{table}[htb]
	\fontsize{8pt}{10pt}\selectfont
	\centering
	\captionsetup{justification=justified, singlelinecheck=false}
	\caption{Summary of metabolic cost across treadmill, complex-terrain, and outdoor overground walking conditions.}
	\label{tab:Metabolic_Summary}
	\renewcommand{\arraystretch}{1.2}
	\resizebox{\columnwidth}{!}{%
		\begin{tabular}{c c c c c c c}
			\toprule[1.2pt]
			\multirow{2}{*}{\textbf{Subject}} & \multicolumn{2}{c}{\textbf{Treadmill}} & \multicolumn{2}{c}{\textbf{Complex}} & \multicolumn{2}{c}{\textbf{Outdoor}} \\
			\cmidrule(lr){2-3}\cmidrule(lr){4-5}\cmidrule(lr){6-7}
			& \textbf{No Assist} & \textbf{Assist} & \textbf{No Assist} & \textbf{Assist} & \textbf{No Assist} & \textbf{Assist} \\
			\midrule[1.2pt]
			P01  & 9.1255 & 8.5950 & 3.5475 & 3.4078 & 3.8208 & 3.6156 \\
			P07  & 8.7975 & 8.0778 & 3.3819 & 3.1293 & 2.2996 & 2.0335 \\
			P10  & 6.7942 & 6.1831 & 3.5852 & 3.2617 & 2.5431 & 2.4612 \\
			P11  & 7.6713 & 6.5534 & 3.6489 & 3.0245 & 3.1465 & 3.0200 \\
			P19  & 7.6817 & 7.0908 & 3.4279 & 3.2807 & 4.2647 & 3.5711 \\
			P24  & 6.3578 & 5.9359 & 3.8371 & 3.7597 & 2.2820 & 2.1588 \\
			P27  & 7.3899 & 6.6958 & 4.0287 & 3.3350 & 3.7287 & 3.3293 \\
			P28  & 7.9111 & 7.3432 & 3.9779 & 3.8398 & 2.6938 & 2.5968 \\
			P30  & 7.5630 & 6.4062 & 3.7066 & 3.4650 & 3.9382 & 3.7681 \\
			\midrule[1.2pt]
			\textbf{Mean} & 7.6991 & 6.9868 & 3.6824 & 3.3893 & 3.1908 & 2.9505 \\
			\textbf{std}  & 0.8178 & 0.8350 & 0.2149 & 0.2534 & 0.7216 & 0.6215 \\
			\bottomrule[1.2pt]
		\end{tabular}%
	}
\end{table}

\begin{figure}
	\centering
	{\includegraphics[width=0.9\columnwidth]{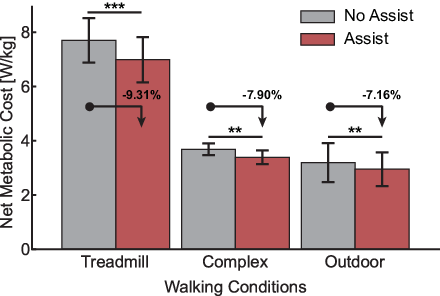}}
	\captionsetup{justification=justified, singlelinecheck=false}
	\caption{Metabolic rate under different walking conditions.}
	\label{fig:metabolic_cost}
\end{figure}

The metabolic costs computed across all walking conditions are summarized in Figure~\ref{fig:metabolic_cost}. 
Under the varying-incline and varying-speed conditions, compared with no assistance, end-to-end control-based assistance led to a significant reduction in metabolic cost of 0.71$\pm$0.24 W/kg (9.31\%$\pm$3.38\%, $p\textless0.05$).
During complex-terrain walking, the metabolic cost under the assistance condition was also significantly reduced by 0.29$\pm$0.21 W/kg (7.90\%$\pm$5.68\%) compared with that under the no-assistance condition ($p\textless0.05$).
With respect to outdoor overground walking, compared with no assistance, end-to-end assistance yielded a significant decrease of 0.24$\pm$0.19 W/kg (7.16\%$\pm$4.57\%) in the metabolic cost ($p\textless0.05$). 
Overall, the end-to-end controller consistently reduced users' metabolic costs across all three walking conditions, suggesting that the K2MUSE dataset has strong potential to support control-oriented applications in wearable robotics.

Compared with the system architecture in \citet{zhang2025modular}, we used longer Bowden cables in our system, which may increase transmission friction and potentially degrade the tracking performance of the assistance profiles. 
To quantify this effect, we computed the RMSE and $R^2$ between the desired hip assistance force profile and the actual force measured by the load cell across the three walking conditions. 
For the treadmill, complex-terrain, and outdoor walking conditions, the RMSE values were 6.54$\pm$0.76 N, 4.95$\pm$1.09 N, and 6.97$\pm$2.18 N, respectively, and the corresponding $R^2$ values were 0.97$\pm$0.03, 0.98$\pm$0.01, and 0.97$\pm$0.08, respectively.
Overall, although the longer cables may introduce additional friction, the tracking error did not increase appreciably relative to that of the shorter-cable configuration in \citet{zhang2025modular}, indicating that this modification does not noticeably compromise the assistance effect.

\section{Usage notes}
\label{UsageNotes}

MATLAB and Python scripts are provided to assist users in visualizing and utilizing data.
These scripts provide instructions for loading the data from the `ProcessedData' folder, generating plots, and performing joint angle estimation via ML models. For additional details on data usage, please refer to the README file.
The metadata in the `*.c3d' files can be directly accessed via \href{https://biomechanical-toolkit.github.io/mokka/index.html}{\textcolor{blue}{the open-source software Mokka}}.
Alternatively, the open-source and cross-platform library \href{https://biomechanical-toolkit.github.io/docs/}{\textcolor{blue}{Biomechanics ToolKit (BTK)}} can also be used to parse `*.c3d' files.

\section{Code availability}
\label{CodeAvailability}

Additional images and videos related to the dataset are available at \href{https://k2muse.github.io/}{\textcolor{blue}{https://k2muse.github.io/}}.
The dataset and codes used to process the data can be found at \href{https://kaggle.com/datasets/98d67c253a7c820668aed0690cae20343481b8f8f8e0dafbe93b0c76d91f0ce6}{\textcolor{blue}{Kaggle -- K2MUSE: A Human Lower Limb Multimodal Dataset}}.
The data can be accessed via MATLAB, and a description of the dataset hierarchy is available at \href{https://k2muse.github.io/datasets/}{\textcolor{blue}{https://k2muse.github.io/datasets/}}.
Joint angle estimation and gait phase classification are implemented via Python.

\begin{itemize}
	\item {\ttfamily scriptProcess.mlx} is used to extract and parse data collected from different devices in the `SourceData' folder and save the data in a unified format.
	\item {\ttfamily scriptPlot.mlx} is used to plot graphs showing the changes in the joint angles and moments during the gait cycle in gait analysis.
	\item {\ttfamily demoRegression.ipynb} is designed to process sEMG, AUS, and kinematic data. It builds ML models for joint angle estimation and analyzes the prediction results.
	\item {\ttfamily demoClassification.ipynb} is designed to process sEMG, AUS, and gait phase labels. It builds ML models for gait phase classification and analyzes the classification results.
	\item {\ttfamily k2muse-transformer-demo} is a transformer-based project for joint angle estimation and gait phase classification.
	\item {\ttfamily control-demo} is a project for deploying trained TCN models with TensorRT. It converts the models into TensorRT engines and performs inference using the TensorRT runtime.
\end{itemize}

\section{Conclusion and future work}

In this paper, we presented the K2MUSE dataset, a large-scale lower-limb locomotion dataset that includes 3D motion trajectories, ground reaction forces, AUS data, and sEMG data. To our knowledge, K2MUSE is the first publicly available dataset to provide these modalities in a synchronized manner across up to 20 ambulation conditions spanning multiple walking speeds, ramp inclines, and representative nonideal acquisition conditions.
The technical validation results confirm the millisecond-level synchronization accuracy and consistent cross-modal coupling of the data, with a dominant coherence peak near 0.90 Hz. Comparisons of the K2MUSE dataset with public datasets show strong agreement (XCOR$\textgreater$0.88). Baseline benchmarks indicate that multimodal fusion enables accurate joint angle estimation (RMSE$\textless$4.5$^\circ$) and gait phase classification (approximately 97.7\% accuracy). We further demonstrated a control-oriented use case with a soft exoskeleton, which achieved consistent reductions in metabolic cost (7.16--9.31\%) across terrains, supporting the utility of the K2MUSE dataset for developing assistance strategies.
Overall, the K2MUSE dataset provides a comprehensive resource for advancing biomechanics, motion-intention recognition, and human--robot interaction in rehabilitation robotics.

In the future, we plan to expand the scope of the K2MUSE dataset by incorporating data collected during additional walking tasks, such as sit-to-stand and sit-to-walk transitions \citep{Huo_TRO}, as well as data collected in broader walking environments, including outdoor settings and stairs.
While the K2MUSE dataset provides a robust normative baseline, direct transfer to pathological gait is not guaranteed. Neurological impairments and limb loss can introduce pronounced asymmetry, altered muscle coordination, and compensatory movement patterns that differ from those associated with able-bodied locomotion and aging-related adaptations. In future releases, we will prioritize expanding participant diversity to include clinical cohorts and will perform studies on adaptation and transfer across populations.
Furthermore, we will gather walking data of participants while they wear assistive robots, thereby enabling comparative analysis with natural walking patterns. We believe that these efforts will further advance the development of rehabilitation robots, and we look forward to the creation of algorithms based on this dataset that can achieve embodied intelligence in robots in future research.

\begin{acks}
	The authors extend their gratitude to all the volunteers and colleagues who contributed to the experiments for this dataset. The authors also appreciate the development and experimental support provided by the State Key Laboratory of Robotics and Intelligent Systems, Shenyang Institute of Automation, Chinese Academy of Sciences.
\end{acks}

\begin{dci}
The authors declare no potential conflicts of interest with respect to the research, authorship, and/or publication of this article.
\end{dci}

\begin{funding}
	The author(s) disclosed receipt of the following financial support for the research, authorship, and/or publication of this article: This work was supported in part by the National Natural Science Foundation of China under Grants 62473361, 62333007, and U22A2067; in part by the Natural Science Foundation of Liaoning Province under Grant 2025JH6/101000028; in part by the Fundamental Research Project of SIA under Grant 2024JC1K01; and in part by the China Postdoctoral Science Foundation under Grant 2024M753412.
\end{funding}

\begin{orc}
	Jiwei Li~\orcidlink{0009-0000-2297-8278}	\href{https://orcid.org/0009-0000-2297-8278}{\textcolor{blue}{https://orcid.org/0009-0000-2297-8278}}\\
	Bi Zhang~\orcidlink{0000-0001-8001-002X}
	\href{https://orcid.org/0000-0001-8001-002X}{\textcolor{blue}{https://orcid.org/0000-0001-8001-002X}}\\
	Xiaowei Tan~\orcidlink{0000-0003-0990-0323}
	\href{https://orcid.org/0000-0003-0990-0323}{\textcolor{blue}{https://orcid.org/0000-0003-0990-0323}}\\
	Wanxin Chen~\orcidlink{0000-0001-5930-0209}
	\href{https://orcid.org/0000-0001-5930-0209}{\textcolor{blue}{https://orcid.org/0000-0001-5930-0209}}\\
	Zhaoyuan Liu~\orcidlink{0009-0002-5195-6128}
	\href{https://orcid.org/0009-0002-5195-6128}{\textcolor{blue}{https://orcid.org/0009-0002-5195-6128}}\\
	Juanjuan Zhang~\orcidlink{0000-0002-3833-487X}
	\href{https://orcid.org/0000-0002-3833-487X}{\textcolor{blue}{https://orcid.org/0000-0002-3833-487X}}\\
	Weiguang Huo~\orcidlink{0000-0002-7370-5189}
	\href{https://orcid.org/0000-0002-7370-5189}{\textcolor{blue}{https://orcid.org/0000-0002-7370-5189}}\\	
	Jian Huang~\orcidlink{0000-0002-6267-8824}
	\href{https://orcid.org/0000-0002-6267-8824}{\textcolor{blue}{https://orcid.org/0000-0002-6267-8824}}\\
	Lianqing Liu~\orcidlink{0000-0002-2271-5870}
	\href{https://orcid.org/0000-0002-2271-5870}{\textcolor{blue}{https://orcid.org/0000-0002-2271-5870}}\\
	Xingang Zhao~\orcidlink{0000-0001-8194-1870}
	\href{https://orcid.org/0000-0001-8194-1870}{\textcolor{blue}{https://orcid.org/0000-0001-8194-1870}}
\end{orc}

\bibliographystyle{SageH}
\bibliography{References}

\end{document}